\documentclass[10pt,twocolumn,letterpaper]{article}

\usepackage{cvpr}              
\usepackage{comment}
\usepackage{graphicx} 
\usepackage[dvipsnames]{xcolor}

\definecolor{cvprblue}{rgb}{0.21,0.49,0.74}
\usepackage[pagebackref,breaklinks,colorlinks,citecolor=cvprblue]{hyperref}

\def\paperID{15957} 
\def\confName{CVPR}
\def\confYear{2024}

\title{
SMIRK: 3D Facial Expressions through Analysis-by-Neural-Synthesis\\
\vspace{-0.2cm}
}

\newcommand{\modelName}{SMIRK\xspace}
\newcommand{\modelNameLong}{\modelName~(Spatial Modeling for Image-based Reconstruction of Kinesics)}

\newcommand{\norm}[1]{\left\lVert#1\right\rVert}

\newcommand{\shapecoeff}{\boldsymbol{\beta}}

\newcommand{\posecoeff}{\boldsymbol{\theta}_{pose}}
\newcommand{\posecoeffall}{\boldsymbol{\theta}}
\newcommand{\jawcoeff}{\boldsymbol{\theta}_{jaw}}

\newcommand{\expcoeff}{\boldsymbol{\psi}_{expr}}
\newcommand{\eyecoeff}{\boldsymbol{\psi}_{eye}} 
\newcommand{\expcoeffall}{\boldsymbol{\psi}}

\newcommand{\numverts}{n_v}

\newcommand{\landmark}{\textbf{k}}

\newcommand{\cam}{\textbf{c}}

\newcommand{\im}{I}
\newcommand{\imt}{\im'}
\newcommand{\ims}{S}
\newcommand{\rend}{R}
\newcommand{\enc}{E}
\newcommand{\encshape}{\enc_{\shapecoeff}}
\newcommand{\encpose}{\enc_{\posecoeffall}}
\newcommand{\encexp}{\enc_{\expcoeffall}}
\newcommand{\mask}{M}
\newcommand{\trans}{T}

\newcommand{\qheading}[1]{\noindent\textbf{#1}}

\newcommand\blfootnote[1]{%
  \begingroup
  \renewcommand\thefootnote{}\footnote{#1}%
  \addtocounter{footnote}{-1}%
  \endgroup
}

\author{\normalsize{George Retsinas\textsuperscript{1$\dagger$} \quad Panagiotis P. Filntisis\textsuperscript{1$\dagger$}  \quad Radek Daněček\textsuperscript{3} \quad  Victoria F. Abrevaya\textsuperscript{3}} \\ \quad \normalsize{Anastasios Roussos\textsuperscript{4}  \quad Timo Bolkart\textsuperscript{3}\footnote{Now at Google.} \quad Petros Maragos\textsuperscript{1,2}}  \\
\vspace{-0.3cm}
\\
\footnotesize{\textsuperscript{1}Institute of Robotics, Athena Research Center, 15125 Maroussi, Greece} \vspace{-0.1cm}\\
\footnotesize{\textsuperscript{2}School of Electrical \& Computer Engineering, National Technical University of Athens, Greece}\vspace{-0.1cm}\\
\footnotesize{\textsuperscript{3}MPI for Intelligent Systems, Tübingen, Germany}\vspace{-0.1cm}\\
\footnotesize{\textsuperscript{4}Institute of Computer Science (ICS), Foundation for Research \& Technology - Hellas (FORTH), Greece}\vspace{-0.1cm}
}

\begin{document}

\twocolumn[{%
\maketitle

\renewcommand\twocolumn[1][]{#1}
\begin{center}
\vspace{-0.20in}
    \centering
    \captionsetup{type=figure}
    \includegraphics[width=\linewidth]{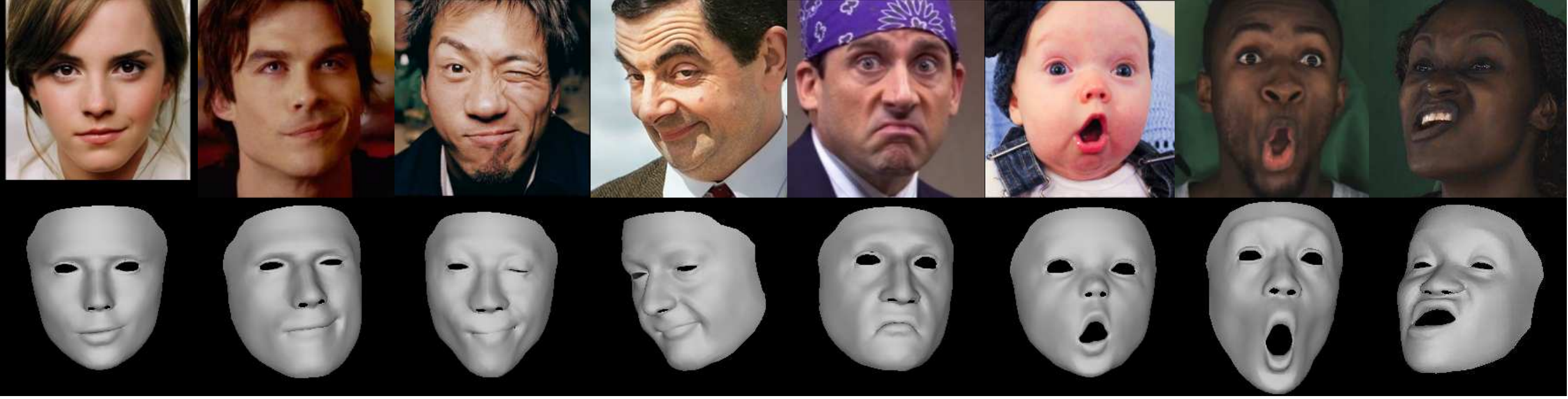}
    \caption{\textbf{\modelName} reconstructs 3D faces from monocular images with facial geometry that faithfully recover extreme, asymmetric, and subtle expressions. Top: images of people with challenging expressions. Bottom: \modelName reconstructions.}
    \label{fig:teaser}
\end{center}
}]

\begin{abstract}
    While existing methods for 3D face reconstruction from in-the-wild images excel at recovering the overall face shape, they commonly miss subtle, extreme, asymmetric, or rarely observed expressions. We improve upon these methods with \modelNameLong, which faithfully reconstructs expressive 3D faces from images. We identify two key limitations in existing methods: shortcomings in their self-supervised training formulation, and a lack of expression diversity in the training images. For training, most methods employ differentiable rendering to compare a predicted face mesh with the input image, along with a plethora of additional loss functions. This differentiable rendering loss not only has to provide supervision to optimize for 3D face geometry, camera, albedo, and lighting, which is an ill-posed optimization problem, but the domain gap between rendering and input image further hinders the learning process. Instead, \modelName replaces the differentiable rendering with a neural rendering module that, given the rendered predicted mesh geometry, and sparsely sampled pixels of the input image, generates a face image. As the neural rendering gets color information from sampled image pixels, supervising with neural rendering-based reconstruction loss can focus solely on the geometry. Further, it enables us to generate images of the input identity with varying expressions while training.
These are then utilized as input to the reconstruction model and used as supervision with ground truth geometry. 
This effectively augments the training data and enhances the generalization for diverse expressions. Our qualitative, quantitative and particularly our perceptual evaluations demonstrate that \modelName achieves the new state-of-the art performance on 
accurate expression reconstruction.
For our method's source code, demo video and more, please visit our project webpage: \url{https://georgeretsi.github.io/smirk/}.

\vspace{-0.4cm}

\end{abstract}

\blfootnote{$\dagger$ Equal contributions. $\qquad$ \text{*} Now at Google.}

\vspace{-.2cm}
\section{Introduction}

Reconstructing 3D faces from single images in-the-wild has been a central goal of computer vision for the last three decades \cite{zollhoefer2018facestar} 
with practical implications in various fields including virtual and augmented reality, entertainment, and telecommunication. 
Commonly, these methods estimate the parameters of a 3D Morphable Model (3DMM) \cite{Brunton2014,egger2020survey}, either through optimization \cite{aldrian2012inverse,Bas2017fitting,Blanz2002,blanzvetter1999,gerig2018morphable,romdhanivetter2005,thies2018face2face} or regression with deep learning \cite{Chang2018_ExpNet,deng2019accurate,genova2018unsupervised,kim2018inverse,Ploumpis2020,richardson2016synthetic,sanyal2019ringnet,tewari17MoFA,AnhTran2017,feng2021learning,danecek2022emoca,filntisis2023spectre}.
%
Due to the lack of large-scale paired 2D-3D data, most learning-based methods follow a self-supervised training scheme using an analysis-by-synthesis approach \cite{blanzvetter1999,tewari17MoFA}.


Although there has been a persistent improvement in the accuracy of identity shape reconstruction, as indicated by established benchmarks \cite{sanyal2019ringnet, feng2021learning}, the majority of works fail 
to capture the full range of facial expressions, including extreme, asymmetric, or subtle movements which are perceptually significant to humans --see e.g. Fig.~\ref{fig:teaser}. 
Recent works addressed this by augmenting the photometric error with image-based perceptual losses based on expert networks for emotion \cite{danecek2022emoca}, lip reading \cite{filntisis2023spectre}, or face recognition \cite{gecer2019ganfit}, or with a GAN-inspired discriminator \cite{otto2023perceptualloss}. 
However, this requires a careful balancing of the different loss terms, and can often produce over-exaggerated facial expressions. 

We argue here that the main problem is the shortcomings of the differentiable rendering loss. 
Jointly optimizing for geometry, camera, appearance, and lighting is an ill-posed optimization problem due to shape-camera \cite{smith2016perspective} and albedo-lighting \cite{egger17phd} ambiguities.
Further the loss is negatively impacted by the large domain gap between natural input image and the rendering.
The commonly employed Lambertian reflectance model is an over-simplistic approximation of the light-face interaction \cite{egger2020survey}, and it is insufficient to account for hard self-shadows, unusual illumination environments, highly reflective skin, and differences in camera color patterns. 
This, in turn, can result in sub-optimal reconstructions by providing incorrect guidance during training. 
In this work, we introduce a simple but effective analysis-by-neural-synthesis supervision to improve the perceived quality of the reconstructed expressions.
For this, we replace the differentiable rendering step of self-supervised approaches with an image-to-image translator based on U-Net~\cite{RonnebergerFB15unet}. 
Given a 
monochromatic
rendering of the geometry together with sparsely sampled pixels of the input image, this U-Net generates an image which is then compared to the input image. 
Our key observation is that this neural rendering provides more accurate gradients for the task of expressive 3D face reconstruction.
This approach has two advantages. 
First, by providing the 
rendered predicted mesh without appearance to the generator, 
the system is forced to \textit{rely on the geometry} of the rendered mesh for recreating the input,
leading to more faithful reconstructions. 
Second, the generator 
can create \emph{novel} images, that modify the expression of the input. 
We leverage this while training with an \emph{expression consistency / augmentation} loss.
This renders a mesh of the input identity under a novel expression, 
renders an image with the generator,
project the rendering through the encoder, and penalizes the difference between the augmented and the reconstructed expression parameters.
By employing parameters from complex and extreme expressions captured under controlled laboratory settings, 
the network learns to handle non-typical expressions that are underrepresented in the data, promoting generalization. 
Our extensive experiments demonstrate that \modelName faithfully captures a wide range of facial expressions (Fig.~\ref{fig:teaser}), including challenging cases such as asymmetric and subtle expressions (e.g., smirking). This result is highlighted by the conducted user study, where \modelName significantly outperformed all competing methods.

In summary, our contributions are:
1) A method to faithfully recover expressive 3D faces from an input image.
2) A novel analysis-by-neural-synthesis supervision that improves the quality of the reconstructed expressions.
3) A cycle-based expression consistency loss that augments expressions during training.

\section{Related Work}


Over the past two decades, the field of monocular 3D face reconstruction has witnessed extensive research and development \cite{egger2020survey,zollhoefer2018facestar}. 
Model-free approaches directly regress 3D meshes \cite{deng2020retinaface,feng2018prn,ruan2021sadrnet, dou2017endtoend,alp2017densereg,Jung2021,Sela2017,Szabo2019,Wei2019,Zeng2019_DF2Net,Wu2020} or voxels \cite{Jackson2017}, or adapt a Signed Distance Function \cite{Park2019_DeepSDF, chatziagapi2021sider,yenamandra2021i3dmm} for image fitting. 
These techniques commonly depend on extensive 3D training data, often generated using a 3D face model. 
However, this dependency can constrain their expressiveness
due to limitations inherent to data creation \cite{deng2020retinaface,feng2018prn,alp2017densereg,Jackson2017,Jung2021,ruan2021sadrnet,Wei2019} and
disparities between synthetic and real images \cite{dou2017endtoend,Sela2017,Zeng2019_DF2Net}. 

Many works estimate parameters of established 3D Morphable Models (3DMMs), like BFM \cite{paysan20093d}, FaceWarehouse \cite{cao2013facewarehouse}, or FLAME \cite{li2017flame}. 
This can be achieved using direct optimization procedure in an analysis-by-synthesis framework \cite{aldrian2012inverse, Bas2017fitting,Blanz2002,Koizumi2020_UMDFA,Ploumpis2020,blanzvetter1999,romdhanivetter2005,li2013realtime, cao2014displaced,garrido2016reconstruction,thies2015realtime,thies2018face2face,thies2016facevr,gerig2018morphable}, but this needs to be applied on novel images every time, which is computationally expensive. 
Recent deep learning approaches offer fast and robust estimation of 3DMM parameters, using either supervised 
\cite{AnhTran2017,tran2018extreme,Chang2018_ExpNet,guo2020towards,kim2018inverse,richardson2016synthetic,Zhu2016_3DDFA, zielonka2022mica, zhang2023accurate}
or self-supervised training, 
for which different types of supervision have been proposed and used in combination, with the most important being the following: 
\textbf{a)} 2D landmarks supervision \cite{deng2019accurate,Liu2017,sanyal2019ringnet,tewari17MoFA, tewari2018self,tewari2019fml,feng2021learning,shang2020self,yang2020facescape} is critical for coarse facial geometry and alignment, but is limited by the sparsity and potential inaccuracy of the predicted landmarks, particularly for complex expressions and poses. Methods that rely on dense landmarks \cite{alp2017densereg,wood2022denselandmarks} overcome the sparsity problem but their accuracy is limited by the inherent ambiguity of dense correspondences across different faces. 
\textbf{b)} Photometric constraints \cite{deng2019accurate,genova2018unsupervised, tewari17MoFA,tewari2018self,tewari2019fml,feng2021learning,shang2020self,yang2020facescape} are particularly  effective for facial data, but are susceptible to alignment errors and depend on the quality of the rendered image. 
\textbf{c)} Perceptual losses have been proven beneficial in aligning the output with human perception~\cite{zhang2018unreasonable}. 
Several methods make use of this by applying perceptual features losses of expert networks
for identity recognition
\cite{genova2018unsupervised,feng2021learning, shang2020self, gecer2019ganfit, deng2019accurate}, emotion \cite{danecek2022emoca} or lip articulation \cite{filntisis2023spectre, he2023speech4mesh}, but are hard to balance with other terms and can sometimes produce exaggerated results, particularly in terms of expressions.



%
%
%
%
We explore an alternative approach, where an image-to-image translation model is coupled with a simple photometric error, encouraging more nuanced details to be explained by the geometry.  

Closer to our work are methods that simultaneously train a regressor network and an appearance model to improve the photometric error signal. Booth \etal \cite{booth20173d,booth20183d} employ a 3DMM for shape estimation coupled with a PCA appearance model learned from images in-the-wild. Grecer \etal \cite{gecer2019ganfit} extend this idea by using a GAN to model the facial appearance more effectively. \cite{tran2018nonlinear, tran2019towards,tewari2018self,tewari2019fml,tewari2021learning}  learn non-linear models of shape and expression while training a regressor in a self-supervised manner. Lin \etal \cite{lin2020towards} refine an initial 3DMM texture while training the regressor. 
Several other works learn neural appearance models for faces from large datasets~\cite{gecer2019ganfit, lee2020uncertainty,lattas2020avatarme,luo2021normalized,bai2023ffhq,lattas2023fitme}.
%
%
%
%
In this work, we do not learn a new appearance model, but directly use a generator for better geometry supervision, achieving significantly improved expression estimation.
Also related to this work are approaches that train a conditional generative model that transforms a rendering of a mesh model into a realistic image, e.g.~\cite{ghosh2020gif,DVP,head2head++,doukas2021headgan,papantoniou2022neural,ding2023diffusionrig}. 
While their focus is on controllable image generation, we investigate here how a generator of average capacity can improve supervision for the task of 3D face reconstruction.

\section{Method: Analysis-by-Neural-Synthesis}
\label{sec:method}

\modelName is inspired by recent self-supervised face reconstruction methods \cite{danecek2022emoca, feng2021learning, filntisis2023spectre, zhang2023accurate} that combine an analysis-by-synthesis approach with deep learning.
While the majority of these works produce renderings based on linear statistical models and Lambertian reflectance, \modelName contributes with a novel neural rendering module that bridges the domain gap between the input and the synthesized output. By minimizing this discrepancy, \modelName enables a stronger supervision signal within an analysis-by-synthesis framework. Notably, this means that neural-network based losses such as perceptual~\cite{johnson2016perceptual}, identity~\cite{deng2019accurate,feng2021learning}, or emotion~\cite{danecek2022emoca} can be used to compare the reconstructed and input images without the typical domain-gap problem that is present in most works.

\subsection{Architecture}
\label{subsec:arch}

\qheading{Face Model:}
\modelName employs FLAME \cite{li2017flame} to model the 3D geometry of a face, which generates a mesh of $\numverts=5023$ vertices based on identity $\shapecoeff$ and expression $\expcoeff$ parameters, extended with two blendshapes $\eyecoeff$ to account for eye closure~\cite{zielonka2022mica}, as well as jaw rotation $\jawcoeff$ parameters. Additionally, we consider the rigid pose $\posecoeff$ 
and the orthographic camera parameters~$\cam$.
For brevity, we refer to all expression parameters (i.e $\expcoeff, \eyecoeff$ and $\jawcoeff$) as $\expcoeffall$, and all global transformation parameters (i.e. $\cam$ and $\posecoeff$) as $\posecoeffall$.

\qheading{Encoder:}
The encoder $\enc(.)$ is a deep neural network that takes an image $I$ as input and regresses FLAME parameters. We separate $\enc$ into three different branches, each consisting of a MobilenetV3~\cite{howard2019searching} backbone:
1) $\encexp$, which predicts the expression parameters $\expcoeffall$, 
2) $\encshape$ that predicts the shape parameters $\shapecoeff$, and 
3) $\encpose$ that predicts the global transformation coefficients $\posecoeffall$.
Formally,
\begin{equation}
  \posecoeffall = \encpose(\im), \quad 
  \shapecoeff = \encshape(\im), \quad 
  \expcoeffall = \encexp(\im).
\end{equation}

Since the main focus of this work is on improving \textit{facial expression} reconstruction, we assume at train time that $\encpose$ and $\encshape$ were pre-trained and remain frozen.
Note that unlike previous methods \cite{feng2021learning, danecek2022emoca, filntisis2023spectre}, $\enc$ does not predict albedo parameters since the neural rendering module does not require such explicit information.

\qheading{Neural Renderer:}
The neural renderer is designed to replace traditional graphics-based rendering with an image-to-image convolutional network $\trans$. The key idea here is to provide $\trans$ with an input image where the face is masked out and only a small number of randomly sampled pixels within the mask remain, along with the predicted facial geometry from the encoder $\enc$. By limiting the available relevant information from the input image, $\trans$ is forced to rely on the predicted geometry from $\enc$ to accurately reconstruct it.

Formally, let $\ims = \rend(\posecoeffall, \shapecoeff, \expcoeffall)$ denote the output of the differentiable rasterization step, where $\ims$ is the monochrome rendering of the reconstructed face mesh. The masking function $\mask(\cdot)$ is applied to the input image $\im$, masking out the face and retaining only a small amount of random pixels within the mask. $\mask(\im)$ is then concatenated with $\ims$, and the resulting tensor is passed through the neural renderer $\trans$ to produce a reconstruction of the original image $\imt = \trans(\ims \oplus \mask(\im))$, 
where $\oplus$ denotes concatenation. 
A crucial property of this module is to assist the gradient flow towards the encoder. 
Therefore, we adopt a U-Net architecture \cite{RonnebergerFB15unet, Isola2016pix2pix, zhu2017cyclegan} for $\trans$, since the shortcuts will allow the gradient to flow uninterrupted towards $\enc$ (an ablation study on this can be found in the Suppl. Mat.). 



\begin{figure}
    \centering
    \includegraphics[width=1.0\linewidth]{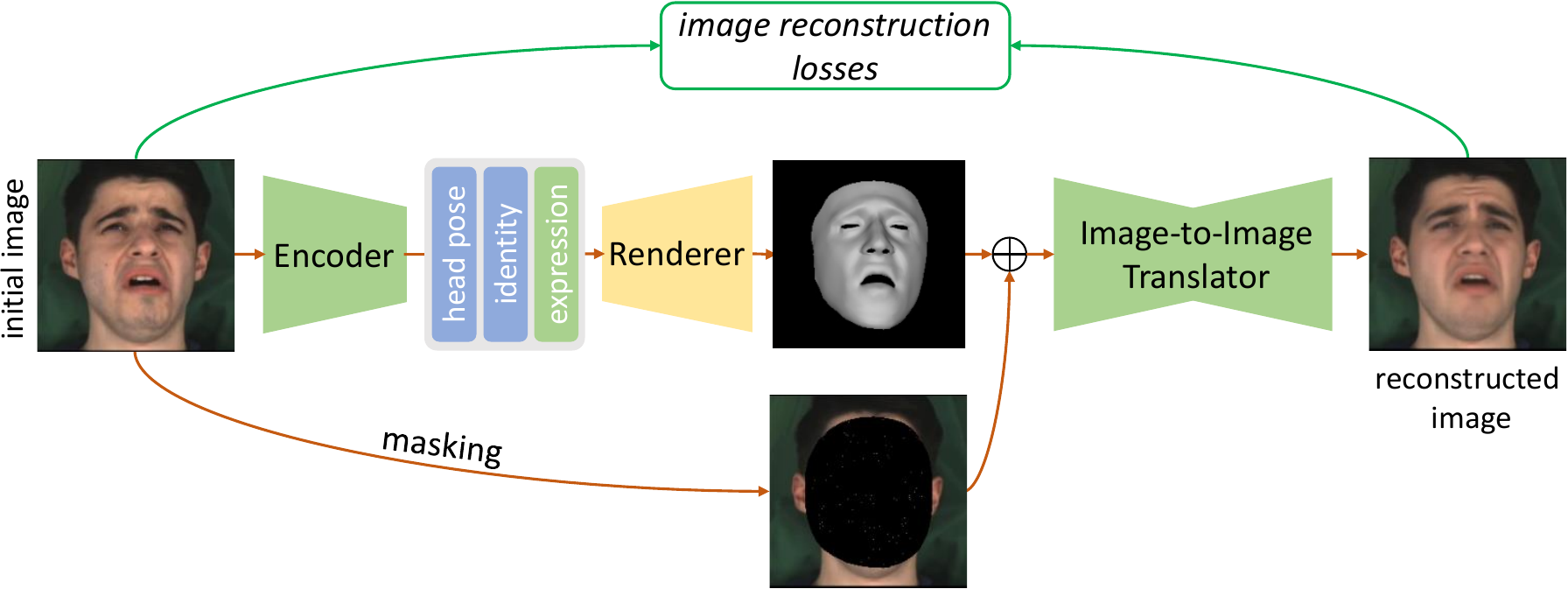}
    \caption{
    \textbf{Reconstruction pass.} An input image is passed to the encoder which regresses FLAME and camera parameters. 
    A 3D shape is reconstructed, rendered with a differentiable rasterizer and finally translated into the output domain with the image translation network. 
    Then, standard self-supervised landmark, photometric and perceptual losses are computed.
    }
    \label{fig:reconstruction}
\end{figure}

\begin{figure}
    \centering
     \includegraphics[width=0.9\linewidth]{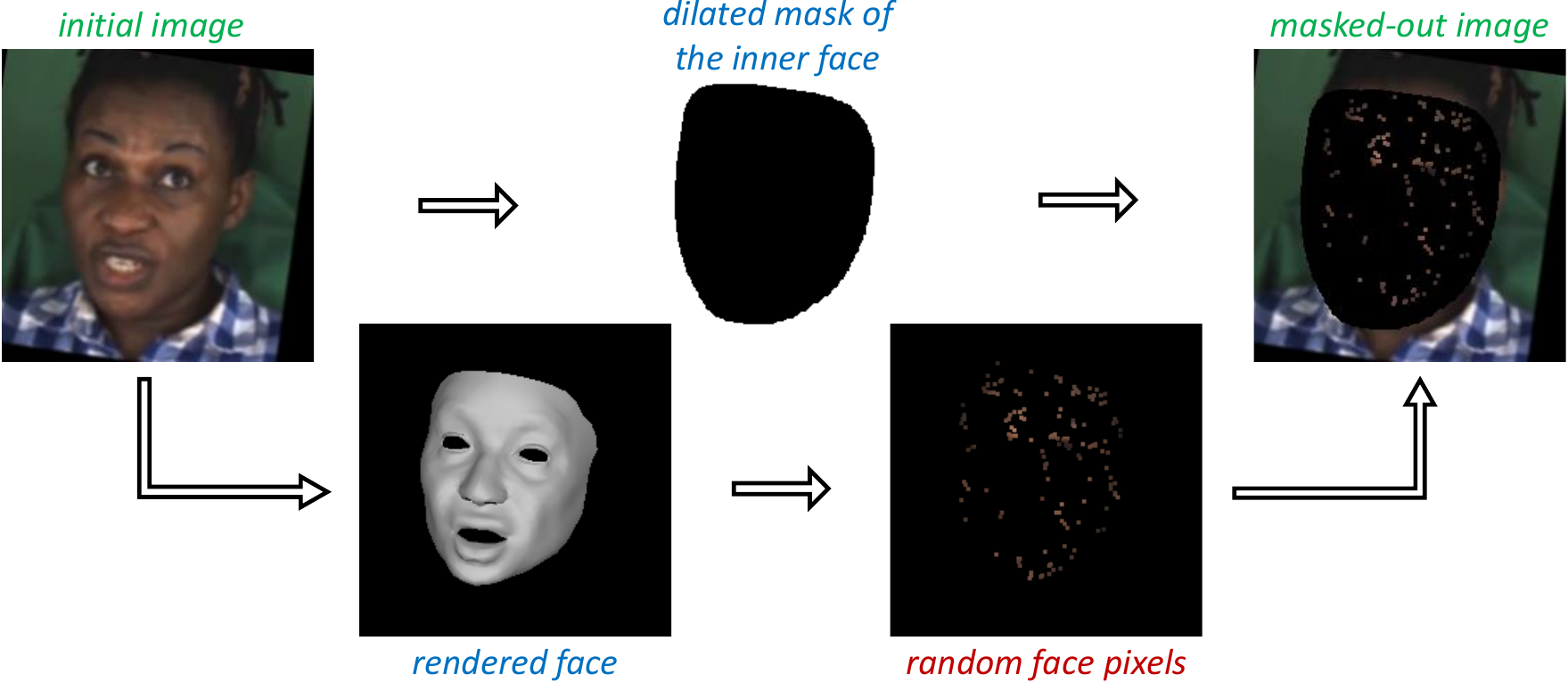} 
    
    \vspace{-0.1cm}
\caption{\textbf{Masking Process.} An input image is masked to obscure the face (upper path), then we sample random pixels to be unmasked (lower path)}.
    \label{fig:masking_procedure}
    \vspace{-0.7cm}
\end{figure}

 

\subsection{Optimization of the SMIRK Components}
\label{subsec:training}
\modelName is supervised with two separate training passes: a \emph{reconstruction} path and an \emph{augmented expression cycle} path. 
We alternate between these passes on each training iteration, optimizing their respective losses. 
We describe each in the following subsections. 


\subsubsection{Reconstruction Path}
In the reconstruction path (Fig.~\ref{fig:reconstruction}), the encoder $\enc$ regresses FLAME parameters from the input image $\im$ and the resulting 3D face is rendered to obtain $\ims$. Next, $\im$ is masked out using the masking function $\mask(.)$, is concatenated with $\ims$, and fed into $\trans$ to obtain a reconstruction of the input image $\imt$.

\textbf{Masking:}
To promote the reliance of $\trans$ on the 3D rendered face for reconstructing $\im$, we need to mask out the face in the input image $I$. We do that by using the convex hull of detected 2D landmarks~\cite{bulat2017far}, dilated so that it fully covers the face. However, without any information of the face interior, training the translator becomes challenging since texture information, such as skin color, facial hair or even accessories (e.g., glasses) are ``distractors" that complicate training. To address this  we randomly sample and retain a small amount of pixels ($1\%$) that are used as guidance for the image reconstruction.
Note that sampling too many pixels makes the reconstruction overly guided and the 3D rendered face does not control the reconstruction output. 
We observed a similar behavior when we tried to randomly mask out blocks of the image, as in ~\cite{he2022autolink}.
The masking process is depicted in Fig.~\ref{fig:masking_procedure}.

\qheading{Loss functions:}
The reconstruction path is supervised with the following loss functions:

\emph{Photometric loss.} This is the L1 error between the input and the output images: $\mathcal{L}_{photo} = \| {\imt} - \im\|_1$.

\emph{VGG loss.} The VGG loss \cite{johnson2016perceptual} has a similar effect to the photometric one, but helps to converge faster in the initial phases of training: $\mathcal{L}_{vgg} = \|\Gamma(\imt) - \Gamma(\im)\|_1$, where $\Gamma(.)$ represents the VGG perceptual encoder.


\emph{Landmark loss.}
The landmark loss, denoted as $L_{lmk} = \sum_{i=1}^{K} \norm{\landmark - \landmark'}_2^2$, measures the $L_2$ norm between the ground-truth 2D facial landmarks detected in the input image ($\landmark$) and the 2D landmarks projected from the predicted 3D mesh ($\landmark'$), summed over $K$ landmarks.

\emph{Expression Regularization.}
We employ an $L_2$ regularization over the expression parameters $L_{reg} = \norm{\expcoeffall}_2^2$, penalizing extreme, unrealistic expressions.

\emph{Emotion Loss.}
Finally, to obtain reconstructions that faithfully capture the emotional content, we employ an emotion loss $\mathcal{L}_{emo}$ based on features extracted from a pretrained emotion recognition network $P_e$, as in EMOCA~\cite{danecek2022emoca}: $\mathcal{L}_{emo} = \| P_e({\imt}) - P_e(\im)\|_2^2 $. To prevent the image translator from adversarially optimizing the emotion loss by perturbing a few pixels, for this loss we keep the image translator $\trans$ ``frozen", optimizing only the expression encoder $\encexp$. Note that unlike EMOCA, our framework ensures that the emotion loss does not suffer from domain gap problems, as the compared images reside in the same space.



\begin{figure}
    \centering
    \includegraphics[width=1.0\linewidth]{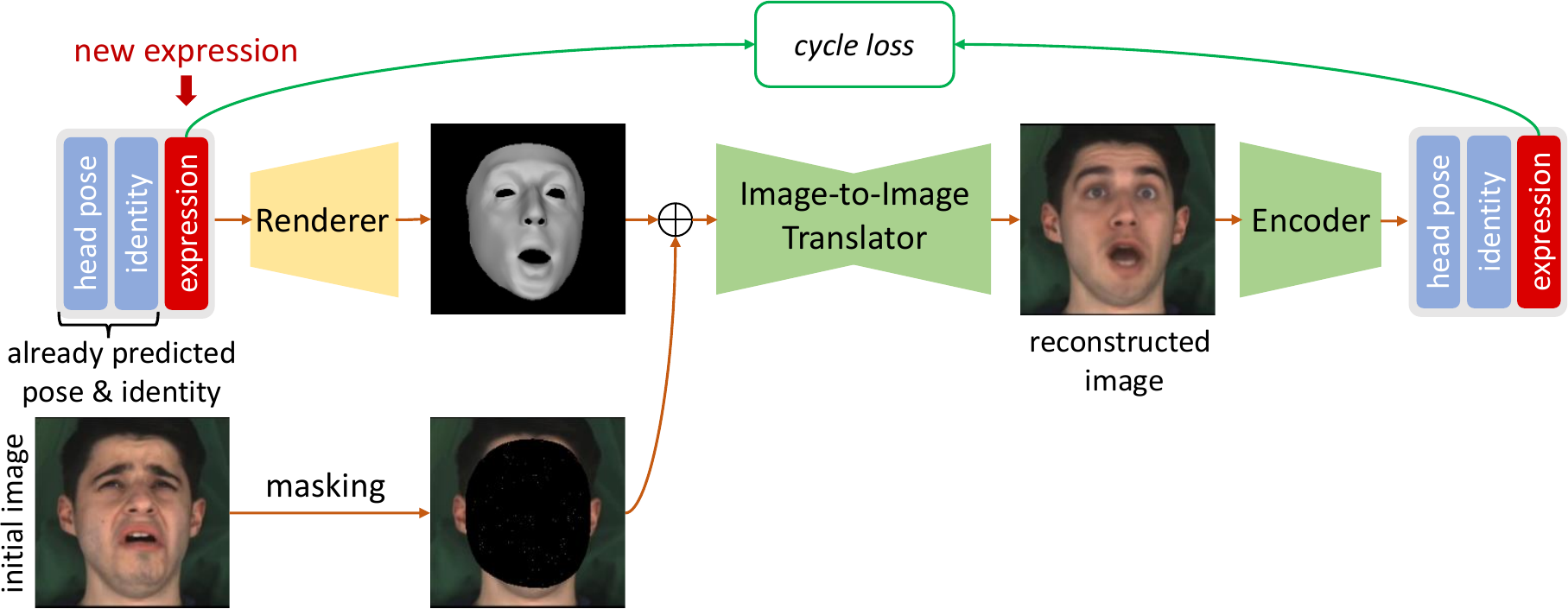}
    \caption{
    \textbf{Augmented cycle pass.} 
    The FLAME expression parameters of an existing reconstruction are modified. The resulting modified face is then rendered using our neural renderer. 
    The rendering is then passed to the face reconstruction encoder to regress the FLAME parameters and a consistency loss between the modified input and reconstructed FLAME parameters is computed. 
    }
    \vspace{-0.2cm}
    \label{fig:cycle_path}
\end{figure}

\subsubsection{Augmented Expression Cycle Path}
\label{sec:cycle_augms}


While the reconstruction path improves 3D reconstruction thanks to the better supervision signal provided by the neural module, 
it is still affected by 
a lack of expression diversity in the training datasets - a problem shared by all previous methods. This means for example that if a more complex lip structure, scarcely seen in the training data, cannot be reproduced fast enough by the encoder, the translator $\trans$ could learn to correlate miss-aligned lip 3D structures and images and thus multiple similar, but distinct, facial expressions will be \emph{collapsed} to a single reconstructed representation. 
Further, this may lead to the translator compensating for the encoder's failures during the joint optimization. 

These issues are addressed with the \emph{augmented expression cycle consistency} path. 
In this path, we start from the predicted set ${\shapecoeff, \expcoeffall, \posecoeffall}$, and replace the original predicted expression $\expcoeffall$ with a new one $\expcoeffall_{aug}$.
We then use the translator $\trans$ to generate a photorealistic image $\imt_{aug}$ which adheres to it. 
This process effectively synthesizes an augmented training pair of $\expcoeffall_{aug}$ and the corresponding output image $\imt_{aug}$. Then, the image is fed into $E$ which should perfectly recover $\expcoeffall_{aug}$. 
A cycle consistency loss can now be directly applied in the expression parameter space of the 3D model, enforcing the predicted expression to be as close as possible to the initial one. 
This concept is illustrated in Fig.~\ref{fig:cycle_path}.

The benefit of this cycle path is two-fold: 1) it reduces over-compensation errors via the consistency loss and 2) it promotes diverse expressions. The latter further helps consistency by avoiding the collapse of neighboring expressions into a single parameter representation. 
Concerning the consistency property, we can distinguish two over-compensating factors. 
First, during the joint optimization of the encoder and the translator, the latter can compensate when the encoder provides erroneous predictions, leading to an overall sub-par reconstruction. 
Second, if we discard the consistency loss, the expression will try to over-compensate erroneous shape/pose, since we assume the shape/pose parameters are predicted from an already trained system and they are not optimized in our framework.
As an example, if the shape parameters do not fully capture an elongated nose, which is an identity characteristic of the person, the expression parameters may compensate this error.
Such behavior is problematic because it entangles expression, shape and pose and adds undesired biases during training.

\qheading{Pixel Transfer:}
The masking process retains a small amount of pixels within the face area.
However, when a new expression is introduced, the previously selected pixels need to be updated and transferred such that they correspond with the vertices of the new expression. 
This operation is referred to as \emph{pixel transfer}, where we sample pixels from the initial image according to a selected set of vertices, we then find the new position of the same vertices for the updated expression, and we assign their position as the new pixel, with the initial pixel value.
This avoids inconsistencies between the underlying structure of the pixels (initial expression) and the new expression, which would hinder realistic reconstructions in the cycle path.

\begin{figure}
    \centering
    \includegraphics[width=1.0\linewidth]{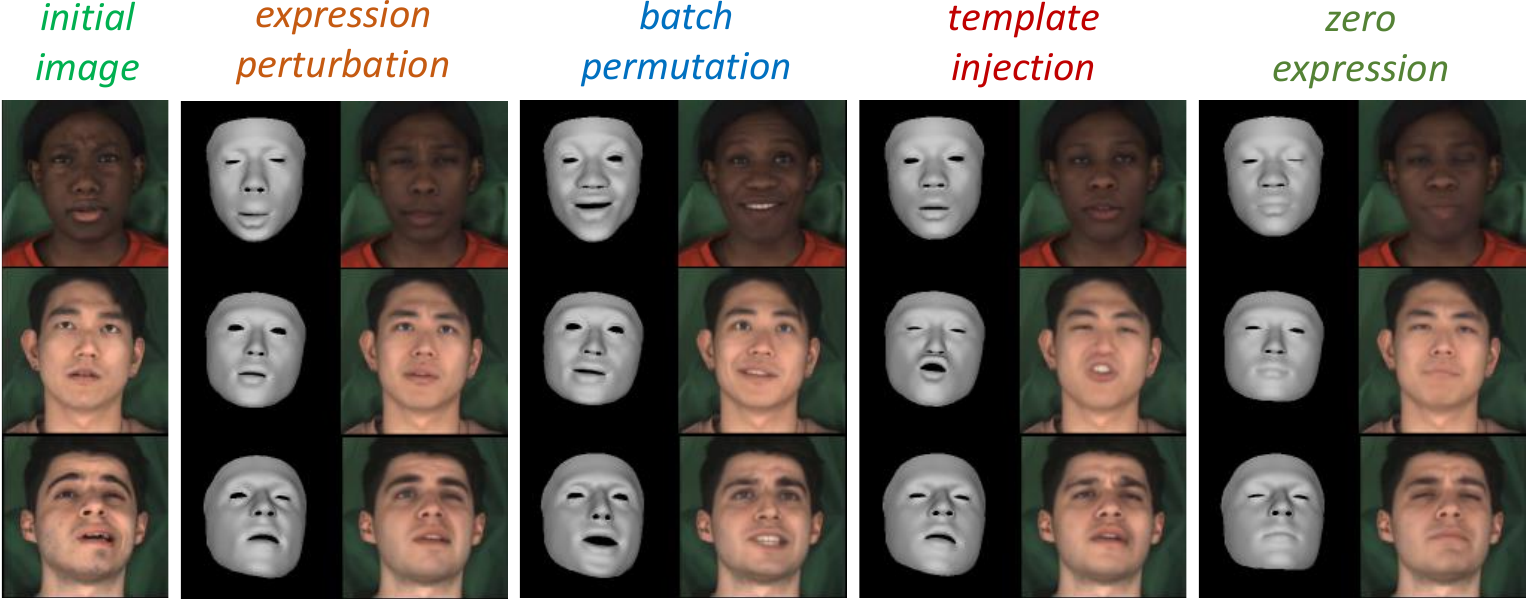}
    \caption{
    \textbf{Neural expression augmentation.} 
    Our neural renderer enables us to modify the expression, generating a new image-3D training pair. 
    We can edit the expression with random noise, permutation from other reconstructions, template injection, or zeroing.
    }
    \vspace{-0.2cm}
    
    \label{fig:cycle_expressions}
\end{figure}

\qheading{Promoting Diverse Expressions:}
Ideally, in this path we also want to promote \emph{high variations in the expression parameter space}, generating shapes (and their corresponding images) with complex, rare and asymmetric expressions that are still plausible.
To effectively augment the cycle path with interesting variations we consider the following augmentations:
\begin{itemize}
\item \textbf{Permutation:} permute the expressions in a batch. 
\item \textbf{Perturbation:} add non-trivial noise to the reconstructed expression parameters.
\item \textbf{Template Injection:} use expression templates of extreme expressions. To obtain such parameters for FLAME we perform direct iterative parameter fitting on the FaMoS~\cite{bolkart2023instant} dataset which depicts multiple subjects perform extreme and asymmetric expressions.
\item \textbf{Zero Expression:} neutral expressions help avoid biasing the system towards complex cases.
\end{itemize}
For all expression augmentations, we simultaneously simulate jaw and eyelid openings/closings, with more aggressive augmentations in the zero-expression case to avoid incompatible blending with intense expressions. Fig.~\ref{fig:cycle_expressions} presents visual examples of all augmentations and the corresponding generated images from $\trans$, showcasing its ability to generate realistic images with notable expression manipulation. 


\qheading{Loss functions:} 

\emph{Expression Consistency.} The expression consistency loss, or cycle loss for brevity, is the mean-squared error between the given augmented expression parameters $\expcoeffall_{aug}$ and the predicted expressions at the end of the cycle path:
\begin{equation}        \mathcal{L}_{exp} = \| \encexp(\trans(\rend(\posecoeffall, \shapecoeff, \expcoeffall_{aug}) \oplus \mask(\im)))
     -  \expcoeffall_{aug}\|_2^2
     \label{eq:expcons}
\end{equation}
The pose/cam and shape parameters are kept as predicted by the initial image, namely $\posecoeffall = \encpose(I)$ and $\shapecoeff = \encshape(I)$.
The internal $\encexp(\im)$ operation, inside the renderer $\rend(\cdot)$, does not allows gradients to flow through and is used as an off-the-self frozen module.   

\emph{Identity Consistency.}
To aid the translator in faithfully reconstructing the identity of the person, we introduce an additional consistency loss similar to Eq. \ref{eq:expcons}, applied to the shape parameters $\shapecoeff$. Note that since the shape encoder $\encshape$ is frozen, the consistency loss only affects the optimization of the translator $\trans$.

\qheading{Alternating Optimization:}
Overall, we alternate between the two passes, aiming to further reduce the effect of the translator compensating for the encoder. 
In more detail, during the augmented cycle pass, we freeze alternatively the encoder and the translator. 
Thus, this pass avoids the joint optimization of the two networks in a single step, acting as a regularizer to the other pass and enforcing consistency.

\section{Results}
We now present objective and subjective evaluations of our method, along with comparisons with recent state of the art.  Additional experimental evaluations and visualizations can be found in our Suppl. Mat. and demo video.

\subsection{Experimental Setup}

\qheading{Training Datasets:}
We use the following datasets for training: FFHQ~\cite{karras2019style}, CelebA~\cite{liu2015faceattributes}, LRS3~\cite{afouras2018lrs3}, and MEAD~\cite{kaisiyuan2020mead}. LRS3 and MEAD are video datasets, and we randomly sample images from each video during training.

\qheading{SOTA Methods:}
We compare with the following recent state-of-the-art methods that have publicly available implementations: DECA~\cite{feng2021learning} and EMOCA v2~\cite{danecek2022emoca,filntisis2023spectre}, which use the FLAME~\cite{li2017flame} model, and Deep3DFace~\cite{deng2019accurate} and FOCUS~\cite{li2021fit}, which use the BFM~\cite{paysan20093d} model.

\qheading{Pretraining:} 
Before the core training stage, all three encoders are pretrained, supervised by two losses - the landmark loss of the reconstruction for pose and expression and the shape predictions of MICA \cite{zielonka2022mica}. After that, $\encshape$ and $\encpose$ remain frozen.

\subsection{Quantitative Evaluations}
It has been consistently reported~\cite{danecek2022emoca,filntisis2023spectre,aldeneh2022towards,garrido2016corrective,mori2012uncanny} that evaluating facial expression reconstruction in terms of geometric metrics is ill-posed. The geometric errors tend to be dominated by the identity face shape and do not correlate well with human perception of facial expressions. Accordingly, we compare our method in a quantitative manner with three experiments:
1) emotion recognition accuracy \cite{danecek2022emoca}, 2) ability of a model to guide a UNet to faithfully reconstruct an input image, and 3) a perceptual user study.

\textbf{Emotion Recognition:}
Following the protocol of~\cite{danecek2022emoca}, we train an MLP to classify eight basic expressions and regress valence and arousal values using AffectNet~\cite{mollahosseini2017affectnet}. We report
Concordance Correlation Coefficient (CCC), root mean square error (RMSE), for both valence (V-) and arousal (A-), and expression classification accuracy (E-ACC). Results are found in Tab.~\ref{tab:emorec}.
As it can be seen, \modelName achieves a higher emotion recognition score compared to most other methods, although falling behind EMOCAv1/2 and Deep3DFace. It is worth noting that, although EMOCA v1 achieves the highest emotion accuracy, it often overexaggerates expressions which helps with emotion recognition. 
EMOCA v2, arguably a more accurate reconstruction model, performs slightly worse. 
Our main model is comparable with Deep3DFace and outperforms DECA and FOCUS. 
We can also train a model that scores better on emotion recognition, by increasing the emotion loss weight. 
However, similarly to what was reported by Daněček et al.~\cite{danecek2022emoca}, this leads to undesirable artifacts. 
We discuss the trade-off between higher emotion recognition scores and reconstruction accuracy in more detail in Sup.Mat.
Notably, even without the emotion loss, the proposed model achieves a decent emotion recognition score, indicating that our reconstruction scheme can adequately capture emotions without the need for explicit perceptual supervision.

\begin{table}[t]
\centering
\resizebox{0.48\textwidth}{!}{
\setlength\tabcolsep{1.5pt}
\begin{tabular}{l|cc|cc|c}
\toprule
        Model &  V-CCC $\uparrow$ &  V-RMSE $\downarrow$ &  A-CCC $\uparrow$ &  A-RMSE $\downarrow$ &  E-ACC $\uparrow$ \\
\midrule
        MGCNet              &  0.69 &   0.35 &  0.58 &   0.34 &     0.60 \\
        3DDFA-v2            &  0.62 &   0.39 &  0.50 &   0.34 &     0.52 \\
        Deep3DFace          &  0.73 &   0.33 &  0.65 &   0.31 &     0.65 \\
        DECA                &  0.69 &   0.36 &  0.58 &   0.33 &     0.59 \\
        FOCUS-CelebA        &  0.69 &   0.35 &  0.54 &   0.33 &     0.58 \\
        EMOCA v1            &  0.77 &   0.31 &  0.68 &   0.30 &     0.68 \\
        EMOCA v2            &  0.76 &   0.33 &  0.66 &   0.30 &     0.66 \\ \hline
        \modelName         &  0.72 &   0.35 &  0.61 &   0.31 &     0.64 \\  
        \modelName w/o emo  &  0.71 &   0.35 &  0.60 &   0.32 &     0.62 \\ 

\bottomrule
\end{tabular}}
\caption{
{{\bf Emotion recognition performance} on the AffectNet test set \cite{mollahosseini2017affectnet}. We follow the same metrics as in \cite{danecek2022emoca}.}
}
\vspace{-0.25 cm}
\label{tab:emorec}
\end{table}

\textbf{Reconstruction Loss:}
\label{sec:recon}
In order to evaluate the faithfulness of a 3D face reconstruction technique, we have devised a protocol based on our analysis-by-neural-synthesis method. Under this protocol, we train a UNet image-to-image translator, but freeze the weights of the encoder so that only the translator is trained. 
The motivation is simple: if the 3D mesh is accurate enough, the reconstruction will be more faithful, due to a one-to-one appearance correspondence. 
For each method (including ours for fairness), we train a UNet for 5 epochs, using the masked image and the rendered 3D geometry as input. 
Finally, we report the $L_1$ reconstruction loss and the 
\emph{VGG}
loss between the reconstructed image and the input image on the test set of AffectNet~\cite{mollahosseini2017affectnet} which features subjects under multiple expressions. The results can be seen in Table~\ref{tab:l1}.
We observe here that using the information for the rendered shape geometry of \modelName, the trained UNet achieves a more faithful reconstruction of the input image when compared to DECA and EMOCAv2. Particularly for EMOCAv2, we observe that although it can capture expressions, the results in many cases do not faithfully represent the input image, leading to an overall worse image reconstruction error. In terms of $L1$ loss, SMIRK is on par with Deep3DFace and FOCUS and has a small improvement in terms of VGG loss. 

\begin{table}[!ht]
\centering
\vspace{-0.0cm}
\resizebox{8.3cm}{!}{
\begin{tabular}{l|l|l|l|l|l}
     & DECA   & EMOCAv2 & FOCUS & Deep3DFace & \modelName    \\\toprule
L1 Loss $\downarrow$ & 0.10 & 0.11 & 0.09 & 0.09 & 0.09 \\
VGG Loss $\downarrow$ & 0.80 & 0.84 &  0.78 & 0.78 & \textbf{0.76} \\
\end{tabular}
}
\vspace{-0.0cm}
\caption{\textbf{Image reconstruction performance} on the AffectNet test set~\cite{mollahosseini2017affectnet}. SMIRK achieves better reconstruction and perceptual scores compared to other methods.}
\vspace{-0.1cm}
\label{tab:l1}
\end{table}



\begin{figure*}
    \centering
    \includegraphics[width=.98\linewidth]{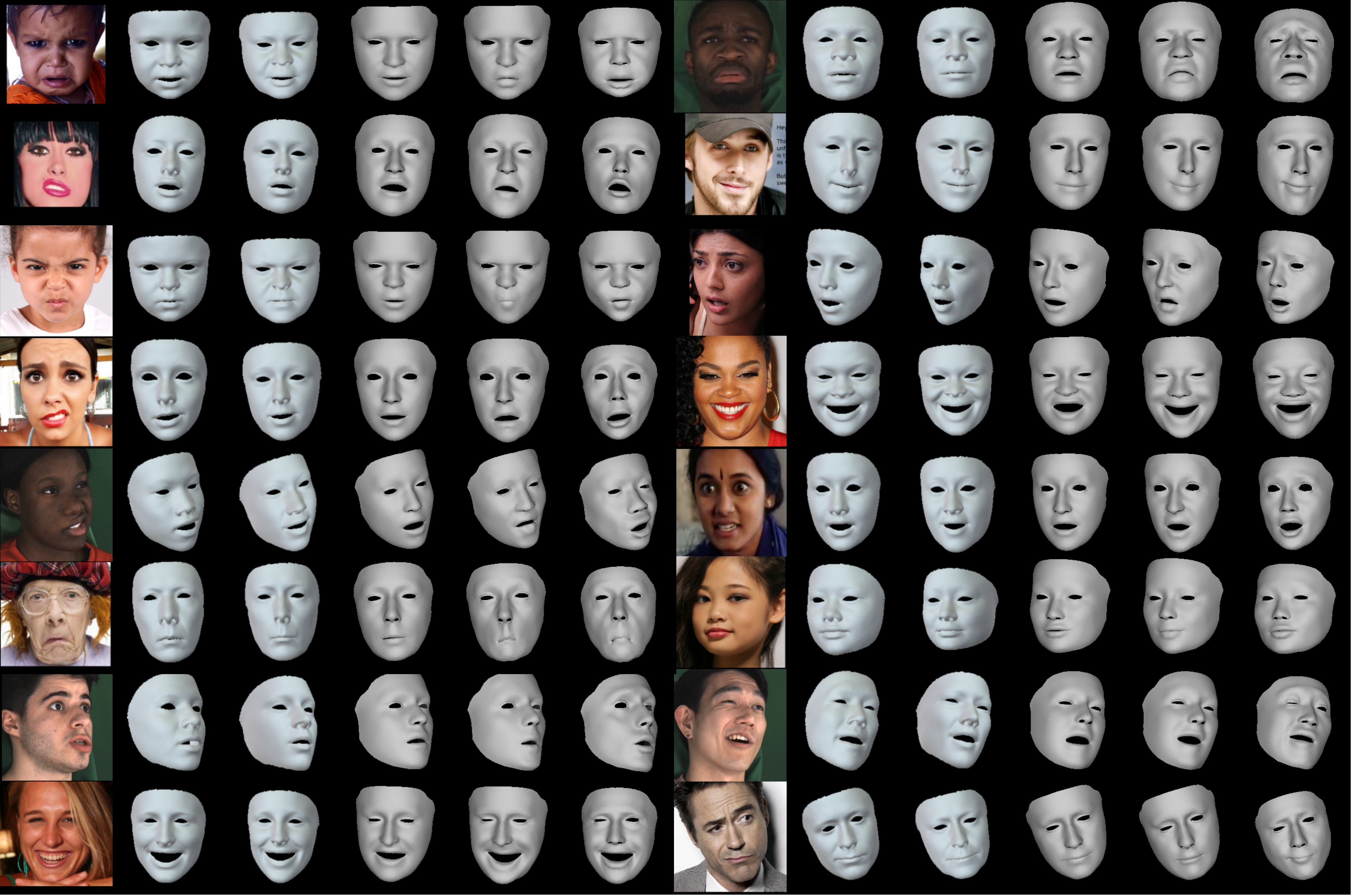}
    \vspace{-.1cm}
    \caption{\textbf{Visual comparison of 3D face reconstruction.} 
    From left to right: Input, Deep3DFaceRecon\cite{deng2019accurate}, FOCUS\cite{li2021fit}, DECA\cite{feng2021learning}, EMOCAv2\cite{danecek2022emoca}, and SMIRK. Many more examples can also be found in the Suppl. Mat. and the demo video in our webpage.}
    \vspace{-.1cm}
\label{fig:visual_examples}
\end{figure*}

\textbf{User Study:} 
Arguably, the perception of the reconstructed facial expressions is the most important aspect in 3D face reconstruction, as it directly influences how well the reconstructed model captures the emotions and nuances of the original face. Considering this, we also designed a user study to assess the perception of the reconstructed facial expressions from human participants. We randomly selected 80 images from the AffectNet \cite{mollahosseini2017affectnet} test set (using the split from \cite{toisoul2021estimation}) and 80 images from our MEAD test set (unseen subjects) and performed 3D face reconstruction with both \modelName and its competitors. To mitigate bias w.r.t. the identity component for the FLAME-based methods, for DECA and EMOCAv2 we used the same identity parameters as our method (which itself was distilled from MICA). In the user study, participants were shown an image of a human face alongside two 3D face reconstructions, either from our method or the others, and were asked to choose the one with the most faithful facial expression representation. The order was randomized for each question, and each user answered a total of 32 questions, equally distributed among the different methods.

A total of 85 users completed the study, and the results in Table~\ref{tab:study1} show that our method was significantly preferred over all competitors, confirming the performance of \modelName in terms of faithful expressive 3D reconstruction. The results were statistically significant (for all pairs, $p<0.01$ with binomial test, adjusted using the Bonferroni method). EMOCAv2, which also uses an emotion loss for expressive 3D reconstruction, was the closest competitor to our method, followed by FOCUS and Deep3D, while DECA was the least selected.


\begin{table}[!ht]
\centering
\resizebox{.85\linewidth}{!}{
\begin{tabular}{l|c|c|c|c}
     & DECA   & EMOCAv2  & Deep3D & FOCUS    \\\toprule
\modelName & \textbf{603}/77& \textbf{461}/219 & \textbf{510}/170  & \textbf{534}/146 \\ 
\end{tabular}
}
\vspace{-0.1cm}
\caption{
\textbf{User study results:} ``a/b" indicates Ours (left) was preferred \textit{a} times, while the competing method was chosen \textit{b} times. SMIRK is overwhelmingly preferred over all other methods.
}
\vspace{-0.0cm}
\label{tab:study1}
\end{table}

\subsection{Visual Examples}
In Fig.~\ref{fig:visual_examples} we present multiple visual comparisons with the four other methods. As it can be visually assessed, our method can more accurately capture the facial expressions across multiple diverse subjects and conditions. Furthermore, the presented methodology can also capture expressions that other methods fail to capture, such as non-symmetric mouth movements, eye closures, and exaggerated expressions.

\subsection{Ablation Studies}


\textbf{Ablation on the effect of landmarks:}
We first assess the effect of the landmark loss. To do that, we calculate for different versions of our model the L1 loss, VGG Loss, and Cycle loss after manipulation of expressions using the same protocol we performed in Sec.~\ref{sec:recon}. Note that this time, we also evaluate performance by considering the \textit{cycle loss}. That is, we also manipulate the predicted expressions, re-generate a new image, and expect that the method can successfully predict the same parameters. We consider three different versions of our model: 1) Protocol 1 - no landmarks loss, 2) Protocol 2 - training some epochs with landmarks loss and then removing it, 3) Protocol 3 - full training with landmarks loss. We present these results in Table \ref{tab:abl1}. 

As we can see, completely omitting landmarks leads to degraded results. However, if we first train for a few epochs with landmarks and then set the loss weight to 0, the model achieves very similar performance with the original model which uses the loss throughout the full training. These results suggest that, in contrast with previous works~\cite{feng2021learning, danecek2022emoca}, the landmarks loss in \modelName acts more as a regularizer during training, helping to guide the model towards good solutions, but in the later stages it may somewhat constrain its flexibility. We plan to explore this balance in more depth in future work.

\begin{table}[!ht]
\small
\centering
\vspace{-0.1cm}
\begin{tabular}{l|l|l|l}
     & L1 Loss $\downarrow$  & VGG Loss $\downarrow$  & Cycle Loss $\downarrow$    \\\toprule
 P1   & 0.111 & 0.757 & 0.588 \\
 P2   & \textbf{0.093} & \textbf{0.713} & \textbf{0.487} \\
 P3  & \textbf{0.093} & 0.714 & 0.544 \\
\end{tabular}
\vspace{-0.1cm}
\caption{\textbf{Ablation study on the effect of landmark loss.} P1: no landmark loss, P2: landmark loss removed after a few epochs, P3: landmark loss throughout whole training.}
\vspace{-0.15 cm}
\label{tab:abl1}
\end{table}

\begin{figure}
    \centering
    \includegraphics[width=0.6\linewidth]{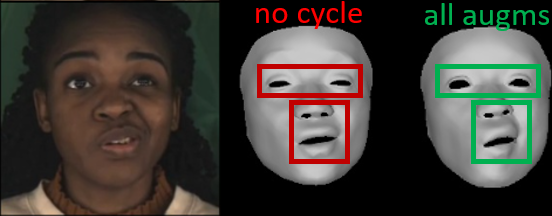}
    \caption{\textbf{Impact of cycle augmentations}. From left to right: input image, no cycle loss, cycle loss with all augmentations.}
    \vspace{-0.2cm}
    \label{fig:cycle}
\end{figure}

\begin{figure}[t]
    \centering
    \includegraphics[width=1.0\linewidth]{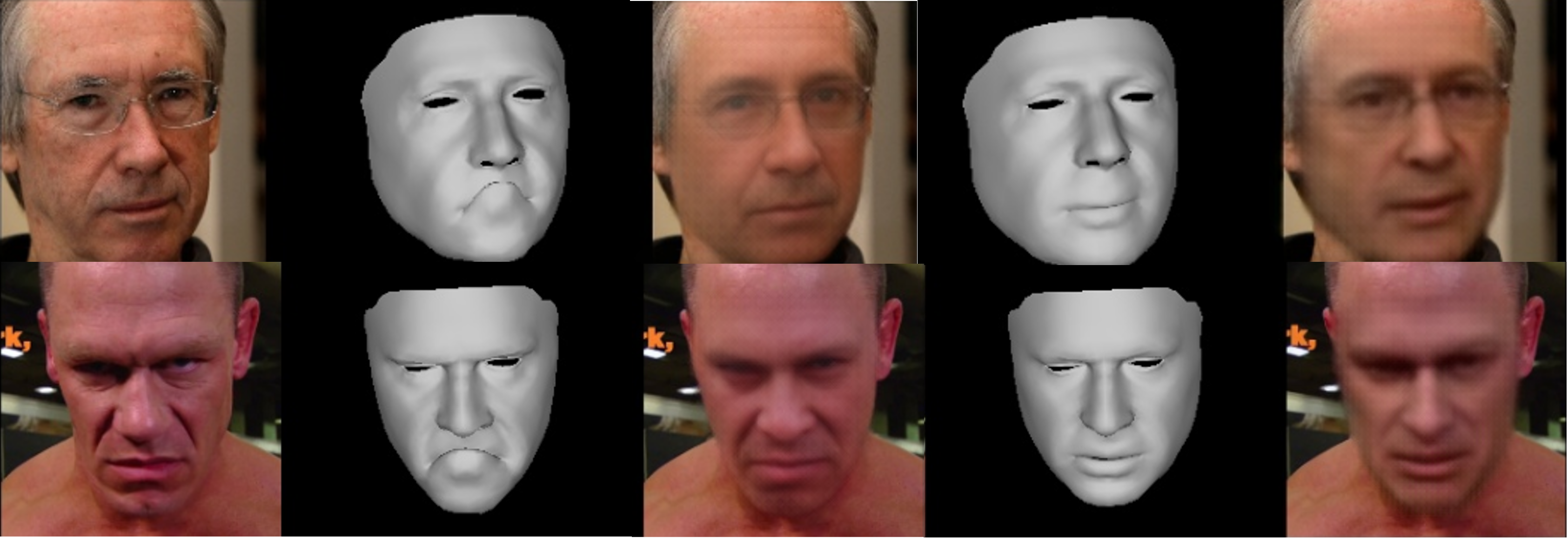}
    \caption{\textbf{Impact of the Cycle Path. }Artifacts can appear when not training with the cycle path. From left to right: input image, 3D reconstruction and image reconstruction \emph{without} cycle path, 3D reconstruction and image reconstruction \emph{with} cycle path.}
    \vspace{-0.25cm}
    \label{fig:artifacts_cycle}
\end{figure}

\vspace{-0.2cm}
\textbf{Impact of Cycle Path:}
Here we also present examples on how the cycle path affects the reconstruction performance. First, we show an example result in Fig.~\ref{fig:cycle}, where we see that using the proposed augmentations provides more detailed expressions. 
For example, template injection augmentation considerably helps the reconstruction of the mouth structure.  Secondly, we have also observed that the cycle path makes the model more robust, especially w.r.t. mouth closures (e.g. zero jaw opening).  We show such indicative cases in Figure~\ref{fig:artifacts_cycle}.
Such artifacts can be seen when using the no-cycle variant, acting as a visual confirmation of the aforementioned numerical results.
Here, the mouth is not properly closed in the 3D reconstructed face, since it was miss-corresponded to a properly closed mouth in the image reconstruction space. The cycle path can solve such instances by providing tweaked expressions that are enforced to be recognized correctly, 
avoiding ``misalignments" between expected expressions and reconstructed images.

\subsection{Limitations}
Despite the effectiveness of \modelName, there are limitations to be addressed. It is sensitive to occlusions, as the training datasets do not include them, and assumes more intense expressions when parts are missing instead of extrapolating from available information. In addition, \modelName has been trained on single images, and the temporal aspect is not yet explored. Also note that while \modelName does not need to predict albedo and lighting, this can be limiting for specific applications in 3D facial animation and video editing. Please refer to the Suppl. Mat. for a more detailed discussion.

\section{Conclusion}
We have presented SMIRK, a new paradigm for accurate expressive 3D face reconstruction from images. Instead of the traditional graphics-based approach for self-supervision which is commonly used for monocular 3D face reconstruction in-the-wild, SMIRK employs a neural image-to-image translator model, which learns to reconstruct the input face image given the rendered predicted facial geometry. Our extensive experimental results show that SMIRK outperforms previous methods and can faithfully reconstruct expressive 3D faces, including challenging complex expressions such as asymmetries, and subtle expressions such as smirking. 

\section*{Acknowledgments}
This research work was supported by the project “Applied Research for Autonomous Robotic Systems” (MIS 5200632) which is implemented within the framework of the National Recovery and Resilience Plan “Greece 2.0” (Measure: 16618- Basic and
Applied Research) and is funded by the European Union- NextGenerationEU.

{
    \small
    \bibliographystyle{ieeenat_fullname}
    \bibliography{main}
}
\clearpage

\section*{\textbf{Supplementary Material}}

\maketitle
\appendix


This supplementary material provides additional details and results for SMIRK. Section \ref{sec:impl} describes the architectural choices and training details. In Section \ref{sec:add_quant}, we provide further quantitative evaluations, and Section \ref{sec:add_abl} presents an extended set of ablation studies to better understand the impact of various components and design decisions. Finally, in Section \ref{sec:limitations}, we discuss the limitations of SMIRK and explore potential future research directions, and Section \ref{sec:add_qual} showcases more qualitative results.

\section{Implementation Details}
\label{sec:impl}
We describe here the implementation details of various subcomponents of the proposed method. 
For more information we refer to our method's source code and demo video: \url{https://georgeretsi.github.io/smirk/}.
\\

\subsection{Image-to-Image Translator}

\begin{figure}
    \centering
    \includegraphics[width=1.05\linewidth]{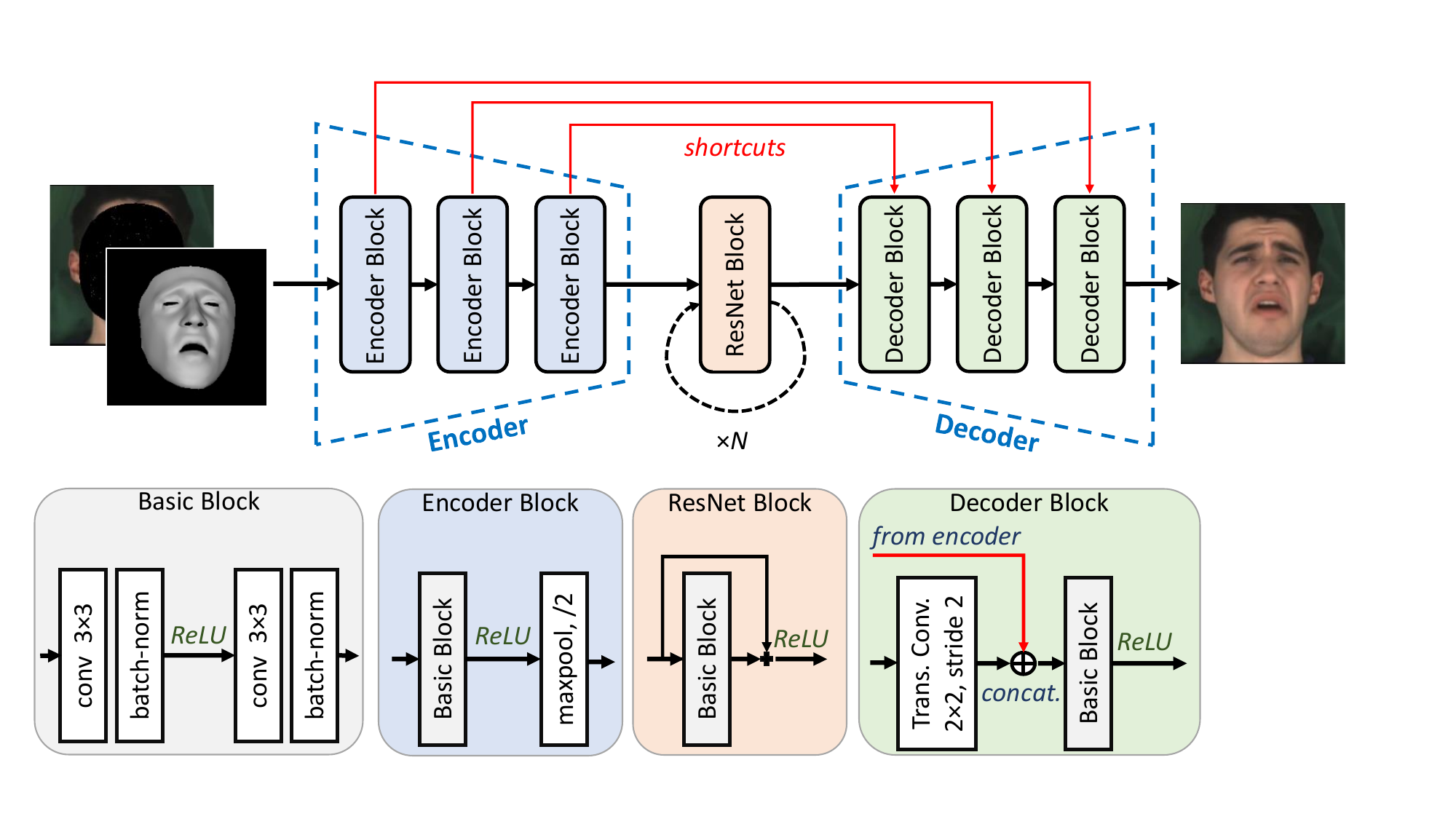}
    \caption{\textbf{Architectural Overview of the Image-to-Image Translator.} The encoder, which consists of 3 encoder blocks, downscales ($/8$) the initial input into a feature tensor map of size $H/8 \times W/8 \times 512$. This feature map is further processed through a set of residual blocks. The image is then reconstructed through the decoder, which consists of 3 decoder blocks. These decoder blocks upscale the feature maps using transposed convolutions, concatenate the resulting feature map with the respective map from the encoder phase using shortcut connections, and process the output with typical convolution operations (Basic Block).}
    \label{fig:generator}
\end{figure}

One important component in the proposed pipeline is the \emph{Image-to-Image Translator}, which relies on UNet architecture~\cite{RonnebergerFB15unet}.
Figure~\ref{fig:generator} depicts this module and all its sub-components.
In more detail, our implementation comprises the typical encoder and decoder convolutional parts, connected with shortcut paths, as shown in Fig.~\ref{fig:generator}. 
Additionally, between the encoder and the decoder, we used a set of residual layers to further process the encoder output. 
The core feature of this module is the shortcut connections, either as residual connections or as UNet connections, that allow the gradients to be easily propagated through the entire network. 
As mentioned before, this image-to-image translation operation should be an appearance-first model, since the geometry of the face is given through the rendered 3D face and the main functionality of the translator resides in inpainting the missing texture. 
We validate the importance of shortcut connections in the ablation study of Sec.~\ref{sec:generator_ablation}.

\subsection{Transfer Pixels in Cycle Path}

One simple, yet effective, component of the augmented cycle path is the \emph{transfer pixel} operation. 
In the cycle path we have a new tweaked expression and thus the facial points that we have selected from the initial image correspond to translated points in the new augmented image. 
If we keep the pixel locations as they are, from the initial image, inconsistencies will arise. 
For example, a pixel that corresponds to the lips in the initial image may correspond to the mouth interior in the tweaked expression. 

Given an initial expression and the new imposed expression, we know the difference between the two corresponding face geometries. 
In other words, if we select a pixel that corresponds to a facial point at the initial image, we can calculate the displacement vector that maps it to the new pixel location of the same facial point at the image with the tweaked expression. In this way, we can sample facial locations that are consistent.  
This observation is the core of this functionality, where we sample some pixels based on the facial geometry of the initial predicted expression, we displace the pixel positions according to the new expression and we assign them the RGB values coming from the initial pixel locations. Formally, given a sparse set of selected pixels with positions $\{x_i\}$ on the initial image $I$, we create an augmented ``guidance" image $I_{aug}$, that samples the interior of the new face, using the displacement vectors $\{d_i\}$ as $I_{aug}(\left\lfloor x_i + d_i \right \rceil) = I(x_i)$ for each $(x_i, d_i)$ pair. Note that image values are RGB triplets.

\subsection{Identity Loss}

Preliminary versions of the SMIRK framework did not include the \emph{transfer pixels} operation. 
Thus we used pixels of the initial un-tweaked image as guidance in the cycle path of different expressions. 
This introduced an inconsistency between reconstruction and cycle path and cycle image reconstruction were non-realistic, following only the rendered expression.
To address this we used an off-the-self perceptual identity loss, implemented via a Resnet50 model pretrained on the VGG-Face2 dataset~\cite{cao2018vggface2,feng2021learning}.

Nonetheless, for the final SMIRK version, where we use the transfer pixels option, the aforementioned issue is minimized.  Instead, we use a \emph{structural} identity loss. As discussed in the main manuscript, this loss uses the frozen shape encoder $\encshape(\im)$ to enforce a structural shape consistency by minimizing the $L_2$ distance between the predicted shape and the original shape. 
This loss acts only on the image-to-image translator $\trans$ and tries to generate accurate image reconstruction by promoting decoupling of the shape/expression parameters.

\subsection{Template Injection}
\begin{figure}
    \centering
    \includegraphics[width=1.0\linewidth]{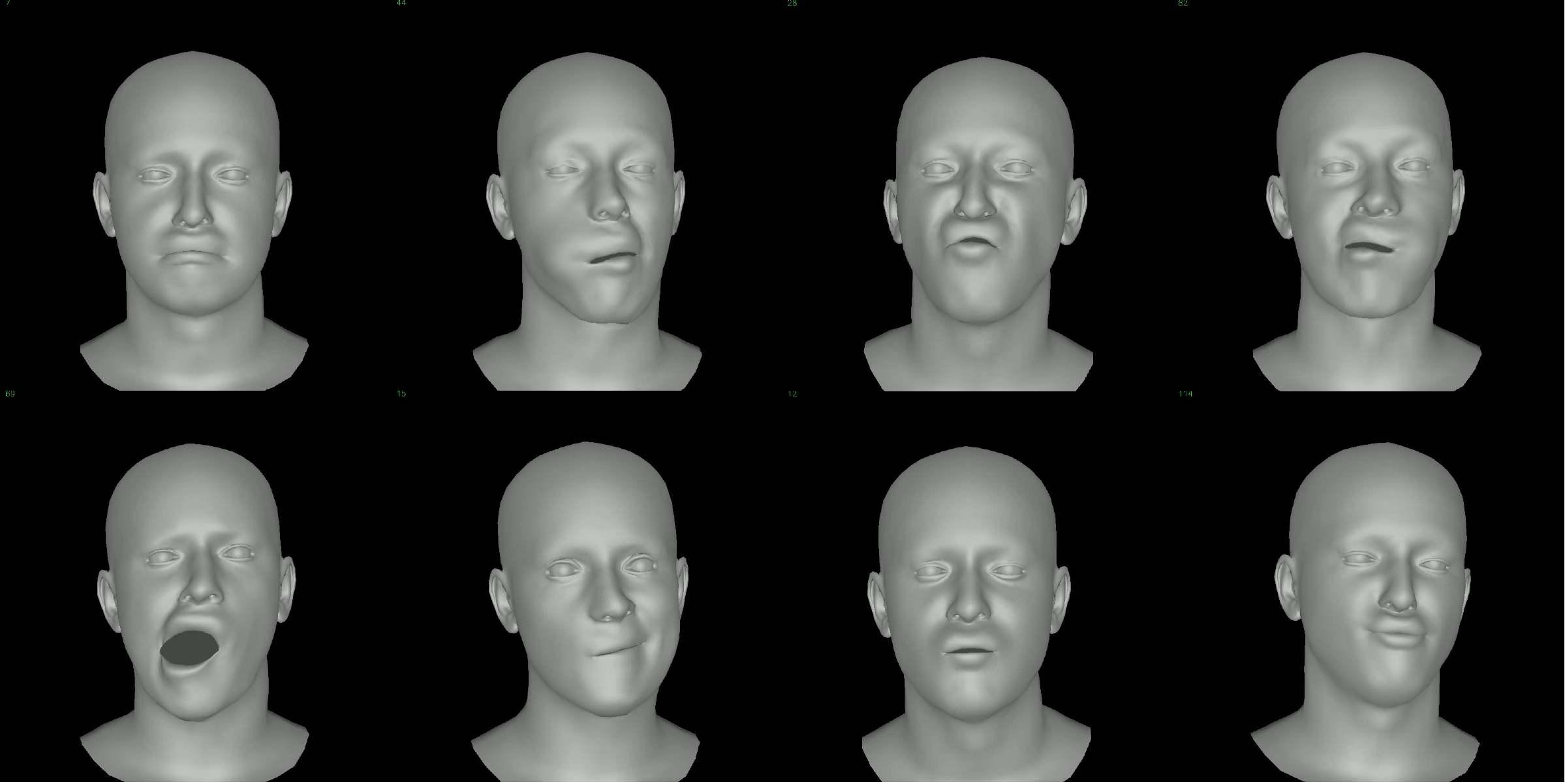}
    \caption{Examples of expression templates used in the cycle path.}
    \label{fig:famos}
\end{figure}

In order to acquire templates (i.e., expression parameters) that correspond to specific, rarely-encountered expressions, we have performed direct iterative parameter fitting on the FaMoS~\cite{bolkart2023instant} dataset. More specifically, we fitted pose and expression parameters of FLAME to the following sequences of the dataset from 70 random subjects, using a sampling stride of 10: 
lips back, rolling lips, mouth side, kissing, high smile, mouth up, mouth middle, mouth down, blow cheeks, cheeks in, jaw, lips up. 
To ensure accurate results 
we used the corresponding neutral template provided for each subject, instead of optimizing the identity parameters.
For parameter fitting we used the  official tensorflow implementation~\cite{tfflame} provided by the authors of FLAME. We present examples of these expression templates using the mean FLAME identity in Figure~\ref{fig:famos}.

\subsection{Model Sizes}
\label{sec:model_sizes}

In this work we aimed for a more lightweight encoder, and hence used MobileNetv3~\cite{howard2019searching} backbones.
Table \ref{tab:comp} reports the number of parameters for SMIRK and the other considered methods. 
As we can see, SMIRK is 14 times smaller than EMOCA/EMOCAv2, and 7 times smaller than other state-of-the-art methods. 
These results further strengthen the superiority of SMIRK, since the considered encoder is of limited capacity. 

\begin{table}[!ht]
\resizebox{8.3cm}{!}{
\begin{tabular}{l|c|c|c|c|c}
             & SMIRK         & DECA  & EMOCAv2 & FOCUS & Deep3d \\ \hline
\# Params & \textbf{3.6M} & 26.8M & 51.4M   & 25.5M & 24.0M 
\end{tabular}
}
\caption{Number of parameters in SMIRK and other SOTA models. SMIRK is 14 times smaller than EMOCA and 7 times smaller than the other methods.}
\label{tab:comp}
\end{table}

\subsection{Training details}
\qheading{Pretraining:} Before training the expression encoder of SMIRK we pretrain all encoders using only landmark losses. During this step a shape regularizer is also added to impose identity shaping with respect to a pre-trained network (MICA~\cite{zielonka2022mica}). The pretraining phase is done for 60,000 iterations using Adam with a learning rate of $5e-4$. 

\qheading{Face Rendering:} FLAME  is a full head model which includes ears, eyeballs, neck, and scalp in the facial mesh. However, in our work we only render the \textit{expressive} part of the 3D model, which is the face. Images of this rendering can be seen in the pipeline figures in the main paper.

\qheading{Training:}
We use the following datasets for training: FFHQ~\cite{karras2019style}, CelebA~\cite{liu2015faceattributes}, LRS3~\cite{afouras2018lrs3}, and MEAD~\cite{kaisiyuan2020mead}. Since LRS3 and MEAD are video datasets, we randomly sample images from each video during training. We train using a batch size of 32, where each batch consists of 50\% images from FFHQ and CelebA to promote in-the-wild reconstruction, 40\% images from MEAD to promote the emotional expressions seen in this dataset, and 10\% images from LRS3, to promote diverse mouth formations during speech. The weights of the losses used for training are $\mathcal{L}_{cycle}=10$, $\mathcal{L}_{lmk}=100$, $\mathcal{L}_{vgg}=10$, $\mathcal{L}_{photo}=1$, $\mathcal{L}_{emo}=1$, $\mathcal{L}_{reg}=1e-3$. 
In the Augmented Expression Cycle Path we augment each predicted sample uniformly with one for each of the augmentations that were described in the main paper. During the core phase we train SMIRK for 250,000 iterations with a learning rate of $1e-3$ and cosine-annealing, restarted at each epoch. 

\paragraph{Landmarks:}
For the landmark loss, like EMOCAv2~\cite{danecek2022emoca}, we use a combination of 92 predicted mediapipe landmarks for the interior of the face and 16 landmarks from FAN\cite{bulat2017far} for the face boundary.

\section{Additional Quantitative Results}
\label{sec:add_quant}
\begin{figure}[h]
    \centering
    \includegraphics[width=.8\linewidth]{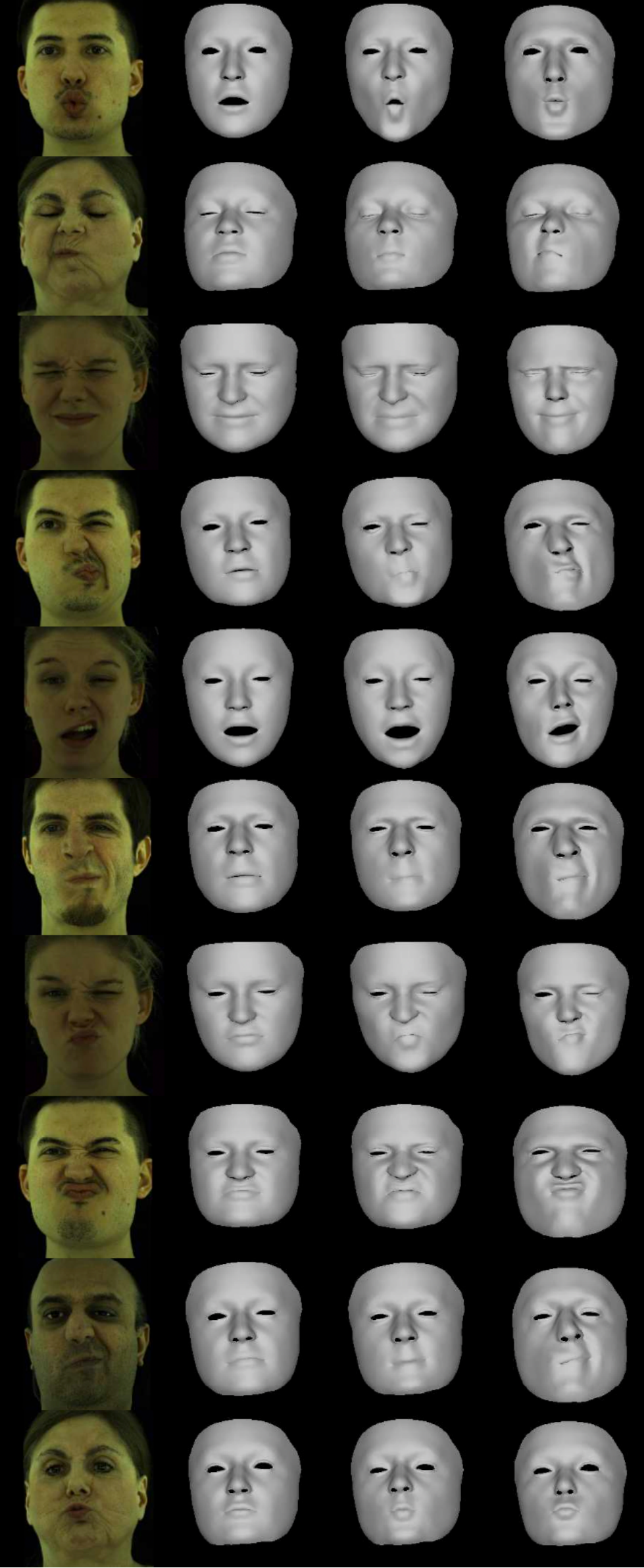}
    \caption{Qualitative comparison of FLAME-based methods on the Multiface dataset. From left to right: Input, DECA\cite{feng2021learning}, EMOCAv2\cite{danecek2022emoca}, SMIRK. \modelName excels in capturing extreme and asymmetric expressions.}
    \label{fig:multiface_results}
\end{figure}

Although as we mentioned in the main text, geometric errors tend to not correlate well with human perception, we also present here the per-vertex errors on the MultiFace \cite{wuu2022multiface} datasets for all FLAME-based methods (which have the same topology).
The MultiFace \cite{wuu2022multiface} v1 dataset consists of 3D scans captured in a multi-camera setup, where subjects where asked to perform various extreme facial expressions. To evaluate the per-vertex error we select the frontal camera subset and select the subjects whose face is fully shown in the image. We use the official test set (``EXP\_ROM07\_Facial\_Expressions''), resulting in a total of 6,324 facial expressions across 5 subjects. In Table~\ref{tab:multiface} we report the mean, median, and max of the ScanToMesh\cite{sanyal2019ringnet} distances between the scans and the predicted mesh surfaces from all FLAME-based methods. Note that the max per-vertex error has been previously reported to correlate better with perceptual quality, compared to the mean that tends to mask inaccurate expressions~\cite{richard2021meshtalk, xing2023codetalker}. As we can see, SMIRK outperforms the other methods on all 3D-reconstruction metrics, and significantly reduces the maximum 3D reconstruction error. Figure~\ref{fig:multiface_results} also shows qualitative comparisons where SMIRK captures significantly more faithfully extreme and asymmetric expressions.

\begin{table}[]
\centering
\begin{tabular}{l|ccc}
        & mean $\downarrow$ & median $\downarrow$ & max $\downarrow$  \\ \toprule
DECA    & 1.40 & 1.12   & 6.8  \\
EMOCAv1 & 1.45 & 1.19   & 6.83 \\
EMOCAv2 & 1.43 & 1.15   & 6.78 \\
SMIRK   & \textbf{1.28} & \textbf{1.05}   & \textbf{5.98} \\
\bottomrule
\end{tabular}
\caption{Per-vertex 3D reconstruction errors (mm) on MultiFace\cite{wuu2022multiface}. SMIRK outperforms other FLAME-based methods.}
\label{tab:multiface}
\end{table}

\section{Additional Ablation Studies}
\label{sec:add_abl}

In this section we explore the impact of several proposed architectural/training options.
\subsection{Impact of Masking}
\begin{figure}
    \centering
    \includegraphics[width=.7\linewidth]{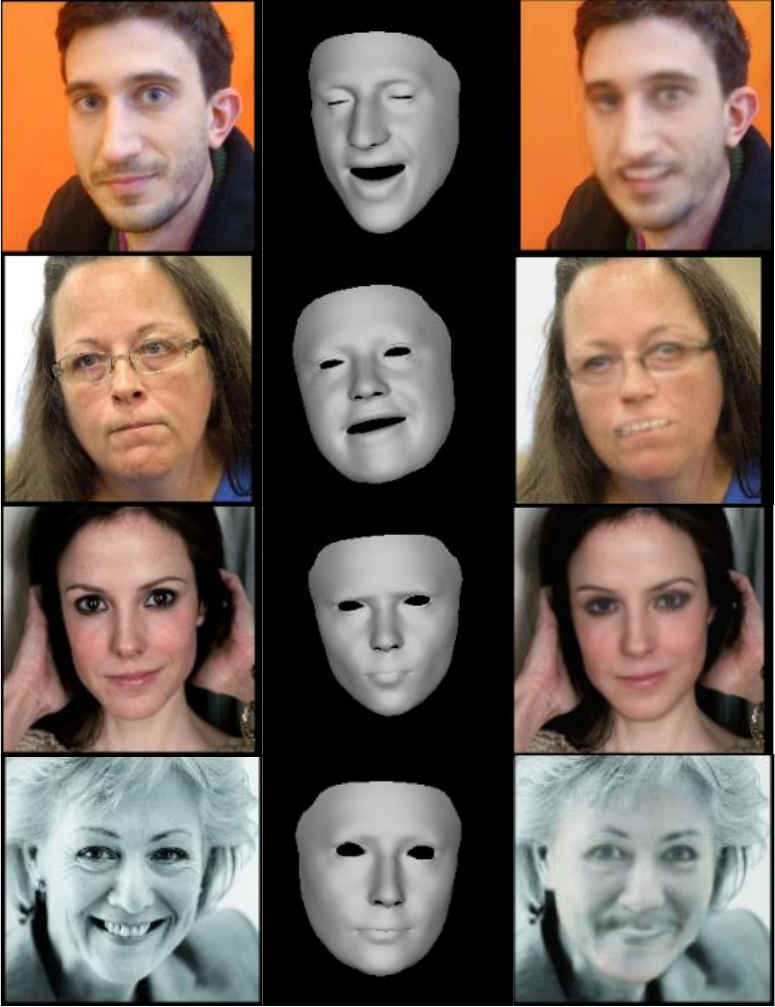}
    \caption{\textbf{Masking with higher percentage of retained pixels.} Left: initial image, Middle: target manipulated expression, Right: reconstructed image. The ratio of pixels to be retained was set to $5\%$ instead of the default $1\%$. We observe that the mouth and eyelid opening/closing cannot be captured adequately, and the emotion is not transferred from the manipulated expression.}
    \label{fig:mr}
\end{figure}

The proposed masking process selects a small number of random pixels inside the face to provide useful texture-related information for the reconstruction of the image.  
We have mentioned that a very small number of pixels is retained, i.e. only $1\%$, since using a higher percentage usually leads to non-realistic inpainting actions.
Such cases are depicted in Fig.~\ref{fig:mr}, where $5\%$ of the pixels are retained.
As we can see, the image reconstruction step struggles to capture different expressions since it relies too much on the selected pixels, with mouth and eyes opening/closing being a major problem. Moreover, emotions cannot be correctly manipulated, as the reconstructed image retains the emotion of the initial image (see e.g. 3rd row of Fig.~\ref{fig:mr}).

\subsection{Impact of Cycle Path}
\label{sec:cycle_ablation}
Here, we perform ablation studies regarding the accuracy of SMIRK with and without the extra augmented expression cycle path, which enables the encoder to see more variations in expressions and further promotes consistency.

\textbf{Image reconstruction}
First, using the protocol in Section 4.2 of the main paper, we train from scratch a UNet image-to-image translator for both the encoders with and without the cycle path. 
We then calculate the reconstruction losses ($L_1$ and VGG losses) on the test set of AffectNet. 
These results can be seen in the first two columns of Table \ref{tab:abl_l1}. 
As we can see, both encoders (with and w/o cycle path) have a very close performance w.r.t. reconstruction metrics, indicating a good correspondence, in average, between the rendered 3D face and the initial image for both alternatives. 

\begin{table}[]
\small
\centering
    \begin{tabular}{l|c|c|l}
        & mean $\downarrow$ & median $\downarrow$ & max $\downarrow$  \\ \toprule
 no cycle path & 1.43 & 1.16 & 6.69 \\
no-injection & 1.32 & 1.07 & 6.08 \\
no-permutation & 1.33 & 1.07 & 6.09 \\
no-zeroing & 1.34 & 1.08 & 6.12 \\
no-random & 1.33 & 1.07 & 6.12 \\
 all augments & $\mathbf{1 . 3 2}$ & $\mathbf{1 . 0 7}$ & $\mathbf{6 . 0 2}$ \\
\bottomrule
\end{tabular}
\caption{Ablation study on the effect of different cycle augmentations on the MultiFace dataset (per-vertex 3D reconstruction errors in mm).}
\label{tab:multiface_ablation}
\end{table}


\begin{table}[h]
\centering
\vspace{-0.1cm}
\resizebox{\linewidth}{!}{
\begin{tabular}{c|cc|cc}
     &  $L_1$ Loss $\downarrow$  & VGG Loss $\downarrow$ & vert $L_1$  $\downarrow$ & vert abs std $\downarrow$\\
\hline
&&&&\\[-2ex]
w/o cycle path & 0.096 & \textbf{0.758} & 0.0130 & 0.0068 \\
w/  cycle path & \textbf{0.095} & 0.762 & \textbf{0.0128}  & \textbf{0.0044}\\
\end{tabular}
}
\vspace{-0.1cm}
\caption{\textbf{Image reconstruction performance} with and without the cycle loss, evaluated on the AffectNet test set~\cite{mollahosseini2017affectnet}. First two columns correspond to the reconstruction metrics, whilst the latter two measure the capability of the generated images to capture changes in expression.}
\label{tab:abl_l1}
\end{table}
\textbf{Capturing small variations}
However, these metrics cannot evaluate the capability of the network to capture small variations in expression.
To do so, we devised a more in-depth ablation study that highlights the adaptability (w.r.t. to expression changes) of a trained translator with and without the cycle path. 
Starting from the inferred FLAME parameters on an input test image (from the test set of AffectNet), we apply $N$ different (minor) augmentations in the expression parameters within a batch, including jaw and eyelids. Then, we use the image-to-image translator to generate a variant of the input face with the new expression and we re-apply the trained encoder to obtain the re-estimated expression parameters, akin to the cycle operation. 
Finally, we calculate:
\begin{itemize}
    \item the $L_1$ norm, dubbed as \emph{vert $L_1$}, between the 3D vertices corresponding to the initial tweaked set of expression parameters, that was used to generate the photorealistic copy, and the 3D vertices corresponding to predicted set of expression parameters.
    We use the comparison on the vertices space to avoid penalizing possible ambiguities in the expression space that the alternative without cycle loss cannot easily discern.
    \item the absolute difference between the standard deviation of the $N$ different copies of each input face, dubbed as \emph{vert abs std}. Again, we calculate this metric between the corresponding vertices.
    This metric indicates how well the encoder can identify minor changes in expressions. 
\end{itemize} 

The aforementioned metrics can be found in the last two columns of Table~\ref{tab:abl_l1}. 
As can be seen, using the cycle path results in similar $L_1$ performance with the non-cycle option (the cycle variant is marginally better), but preserves considerably better the standard deviation between the different image copies. 
The latter is a strong indicator that training with the proposed cycle path helps retaining the variability of the expression parameter space through the translator.

Note that the encoder trained with the cycle path option has seen reconstructed images and used them as input, whilst the encoder train without the cycle path has not, 
Thus, to ensure that these improvements using the cycle path are not fictitious due to a possible distribution shift between the generated images of the two alternatives, we 
re-run the above experiment without tweaking the original expression.
Thus, both encoders are used on the non-altered reconstructed images as sanity check, in essence validating if the generated images in both cases are realistic enough and close to the initial domain. 
In this case both encoders performed equally well ($0.0129$ without cycle path and $0.0128$ with cycle path), 
which shows that no notable domain shift, capable of favoring the one alternative over the other, is evident.

\textbf{Per-vertex reconstruction error}
Finally, we also assess the impact of the cycle path and the different  augmentations in terms of 3D per-vertex reconstruction error. To do this we train separate models for 20 epochs on FFHQ. Results can be found in \ref{tab:multiface_ablation}, for the MultiFace dataset. As we can see, 
best results occur with all augmentations combined, while removing individual augmentations leads to decreased results. Removing the cycle path completely, considerably drops performance.

\begin{figure}[t]
    \centering
    \includegraphics[width=.49\linewidth]{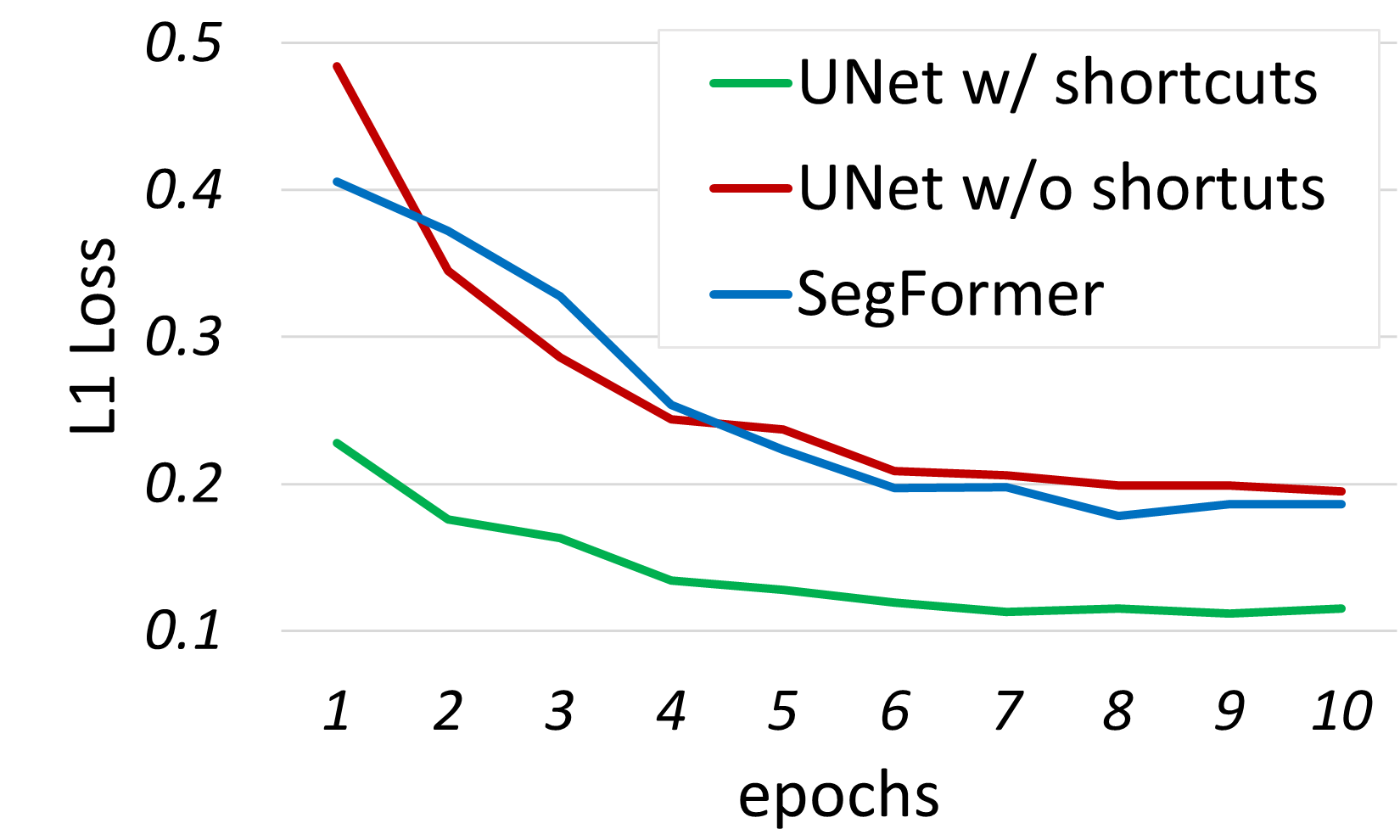}
     \includegraphics[width=.49\linewidth]{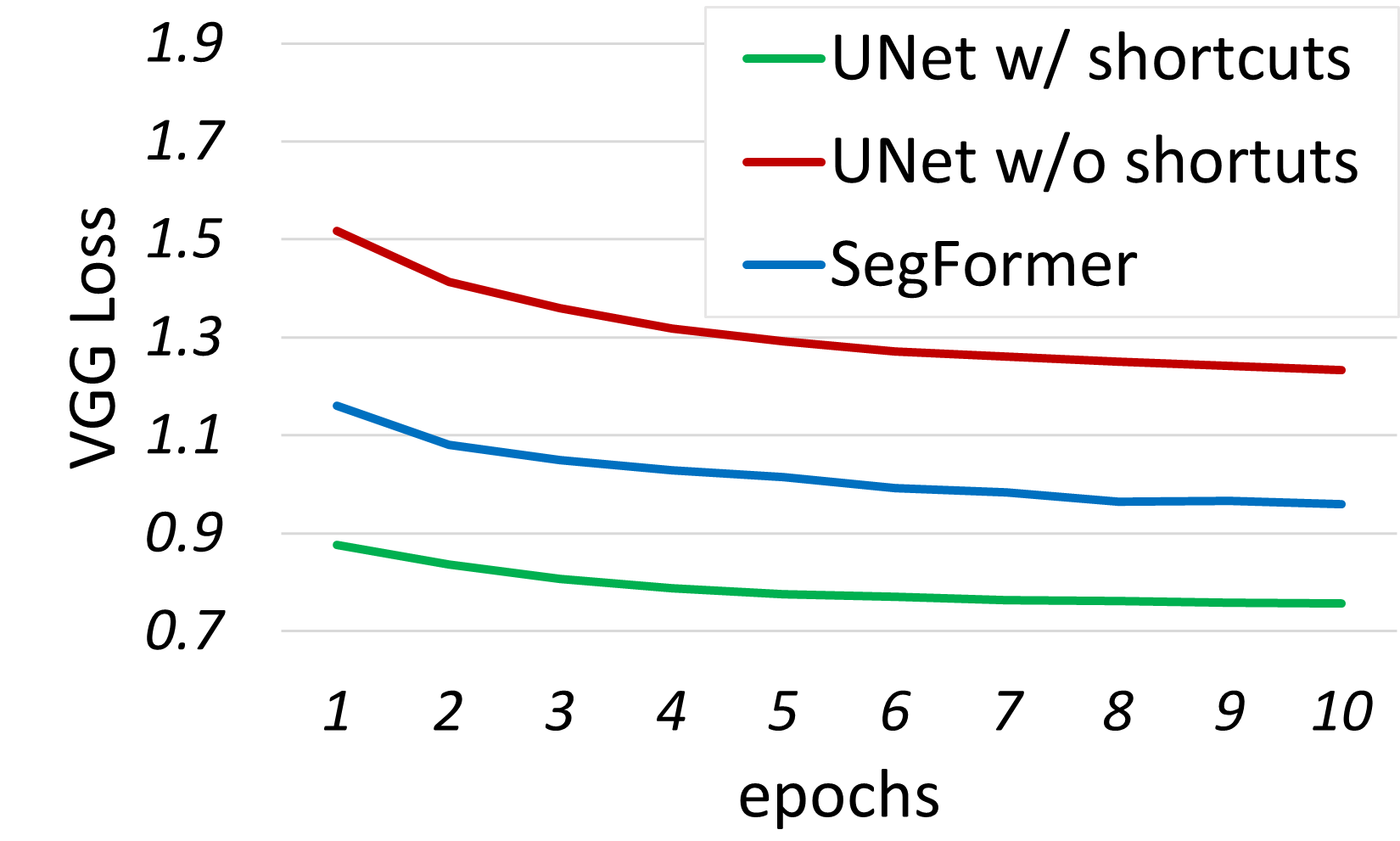}
    \caption{\textbf{Image reconstruction performance for different Translators,} using L1 loss (left) and VGG loss (right).}
    \label{fig:generators_abl}
\end{figure}

\subsection{Impact of Translator's Architecture}
\label{sec:generator_ablation}

One critical property of the Translator, under the proposed framework, is the ``uninterrupted" gradient flow. As we have already described, this is achieved through the shortcut connections of the proposed architecture (as shown in Fig.~\ref{fig:generator}). To validate the importance of these shortcuts connections, we simulate the same architecture (exact same number of parameters) without shortcuts (neither UNet nor residual shortcuts).  
We also consider the transformer-based SegFormer architecture~\cite{xie2021segformer} as an alternative.
The UNet variants have $\sim30$M trainable parameters, while the SegFormer has $\sim85$M parameters.

We trained these three architectural variants with the proposed framework for 10 epochs. The evaluation protocol is the same as in Sec~\ref{sec:cycle_ablation}. The progress of L1 and VGG losses through these 10 epochs is depicted in Figure~\ref{fig:generators_abl}. 
We can observe that the default option with the shortcut connections has a fast convergence to meaningful reconstructions, letting the framework to focus on discovering subtle expression details.
The other alternatives struggle in the first epochs to adapt to the image reconstruction task, which may have a negative impact on the 3D prediction step.  



\begin{table*}[ht]
    \centering
\resizebox{1.\linewidth}{!}{
\begin{tabular}{l|cccc|cccc|c}
\toprule
        emotion weight &  V-PCC $\uparrow$ &  V-CCC $\uparrow$ &  V-RMSE $\downarrow$ &  V-SAGR $\uparrow$ &  A-PCC $\uparrow$ &  A-CCC $\uparrow$ &  A-RMSE $\downarrow$ &  A-SAGR $\uparrow$ &  E-ACC $\uparrow$ \\
\midrule
          0 (w/o emot. loss)   &  0.72 &  0.71 &   0.35 &   0.79 &  0.62 &  0.60 &   0.32 &   0.79 &     0.62\\
         1 (default)       & 0.72 &  0.72 &   0.35 &   0.79 &  0.64 &  0.61 &   0.31 &   0.79 &     0.64 \\
        2        &  0.73 &  0.71 &   0.34 &   0.80 &  0.62 &  0.60 &   0.32 &   0.80 &     0.62 \\
        5        &   0.75 &  0.74 &   0.33 &   0.81 &  0.65 &  0.63 &   0.31 &   0.77 &     0.65\\
        10       &  0.74 &  0.72 &   0.33 &   0.79 &  0.64 &  0.63 &   0.32 &   0.76 &     0.63  \\
\bottomrule
\end{tabular}
}
    \caption{\textbf{Emotion recognition results} for different emotion weights.}
    \label{tab:em_weights}
\end{table*}

\begin{table*}[ht]
\resizebox{\linewidth}{!}{
\begin{tabular}{l|c|c|c|c|c|c|c|c|c}
\toprule
method             & neutral$\uparrow$ & happy$\uparrow$ & sad$\uparrow$  & surprise$\uparrow$ & fear$\uparrow$ & disgust$\uparrow$ & anger$\uparrow$ & contempt$\uparrow$ & avg. (macro)$\uparrow$  \\ 
\midrule
Deep3D             & 0.62    & 0.81  & 0.62 & 0.60     & 0.72 & 0.59    & 0.60  & 0.53     & 0.63 \\
DECA               & 0.45    & 0.76  & 0.51 & 0.50     & 0.69 & 0.64    & 0.59  & 0.54     & 0.58 \\
FOCUS              & 0.50    & 0.75  & 0.47 & 0.65     & 0.51 & 0.61    & 0.56  & 0.58     & 0.58 \\
EMOCA v1           & 0.50    & 0.79  & 0.69 & 0.67     & 0.69 & 0.76    & 0.70  & 0.66     & 0.68 \\
EMOCA v2           & 0.52    & 0.77  & 0.58 & 0.66     & 0.66 & 0.73    & 0.73  & 0.68     & 0.67 \\ \hline
SMIRK w/o em. loss & 0.55    & 0.72  & 0.54 & 0.65     & 0.58 & 0.62    & 0.58  & 0.70     & 0.62 \\
SMIRK $\mathcal{L}_{emo}=1$   & 0.56    & 0.79  & 0.60 & 0.64     & 0.57 & 0.70    & 0.61  & 0.59     & 0.64 \\
SMIRK $\mathcal{L}_{emo}=2$   & 0.44    & 0.74  & 0.59 & 0.65     & 0.61 & 0.59    & 0.66  & 0.69     & 0.62 \\
SMIRK $\mathcal{L}_{emo}=5$   & 0.60    & 0.77  & 0.63 & 0.64     & 0.58 & 0.66    & 0.59  & 0.70     & 0.65 \\
SMIRK $\mathcal{L}_{emo}=10$  & 0.47    & 0.75  & 0.56 & 0.67     & 0.64 & 0.62    & 0.63  & 0.74     & 0.64 \\
\bottomrule
\end{tabular}
}
\caption{\textbf{Accuracy per emotion} for all methods and average (macro).}
\label{tab:per_emotion}
\end{table*}

\begin{figure}[t]
    \centering
    \includegraphics[width=1.0\linewidth]{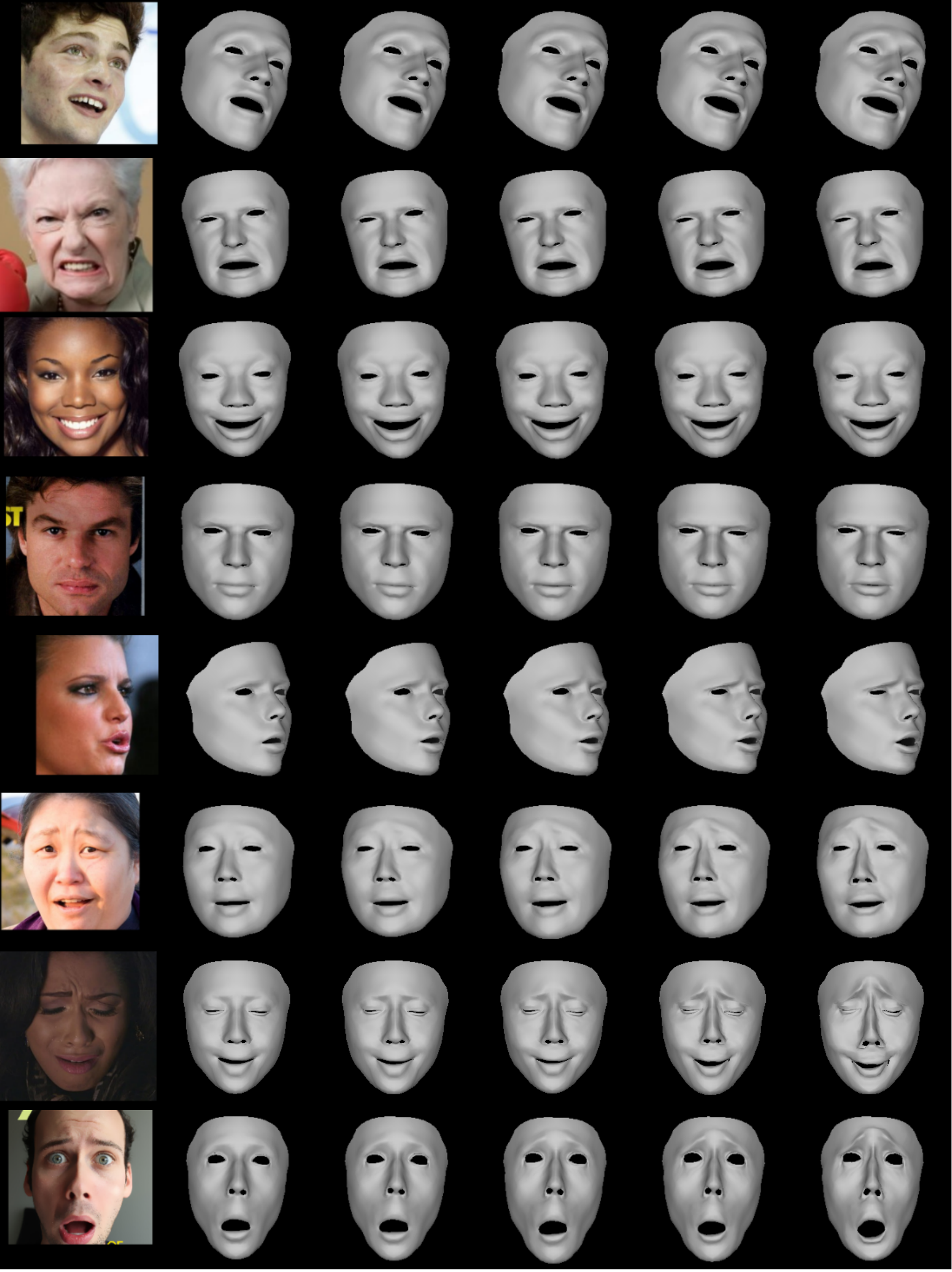}
    \caption{Image results on the effect of emotion loss weight. From left to right, $\mathcal{L}_{emo}=0,1,2,5,10$. We see that in certain cases, higher emotion losses can lead to exaggerated expressions and artifacts. }
    \label{fig:emotion_ablation}
\end{figure}

\subsection{Impact of Emotion Loss}

One of the advantages of the proposed approach is the direct comparison between the reconstructed image and the input image via perceptual losses, without any domain gap involved.
In this work, we considered an extra emotion loss, following EMOCA~\cite{danecek2022emoca}. 
The goal is straightforward: assist the encoder to better capture emotion-related expressions. 

One can tune the contribution of this auxiliary loss through its respective weight, used to calculate the overall loss. 
Using very small values has minor to no impact, while large values cause over-exaggerations of the requested emotions, leading to visually unfaithful 3D reconstructions, as was the case in EMOCA~\cite{danecek2022emoca}. 
The emotion weight was set to $1$, as the default option, after visual inspection for possible expression over-exaggerations. 

Using the protocol of~\cite{danecek2022emoca}, and complementary to the results in Section 4.2 of the main paper we show the effect of different emotion weight in Table~\ref{tab:em_weights}. In addition, more in-depth exploration of the emotion recognition performance is given in Table~\ref{tab:per_emotion}, where we also report per-emotion accuracy, along with the average across emotions, for different emotion weights, as well as the considered SOTA methods. 

We observe that different emotion weights can result in different and non-canonical pertubations in the results, e.g., for emotion loss weight $1$ the accuracy for contempt drops drastically w.r.t. using no emotion, while a similar effect occurs when increase the weight from 5 to 10 for sadness. We also see that the trained MLPs tend to confuse the negative emotion (fear, disgust, anger), and more succesfully predict happiness. Overall, this behavior of the emotion recognition results could be attributed to a possible sensitivity of the trained MLPs combined with the ambiguous nature of emotion classification. In Figure \ref{fig:emotion_ablation} we also show qualitative examples on the effect of emotion loss on images from the AffectNet dataset. We can see that in many cases (rows 1 - 4) the effect of emotion loss weighting is smaller, however for certain emotions san as sadness, the results tend to get very exaggerated as the emotion loss increases. This can result in serious artifacts with higher emotion losses (see last 2 rows).  

\begin{figure}
    \centering
    \includegraphics[width=1.0\linewidth]{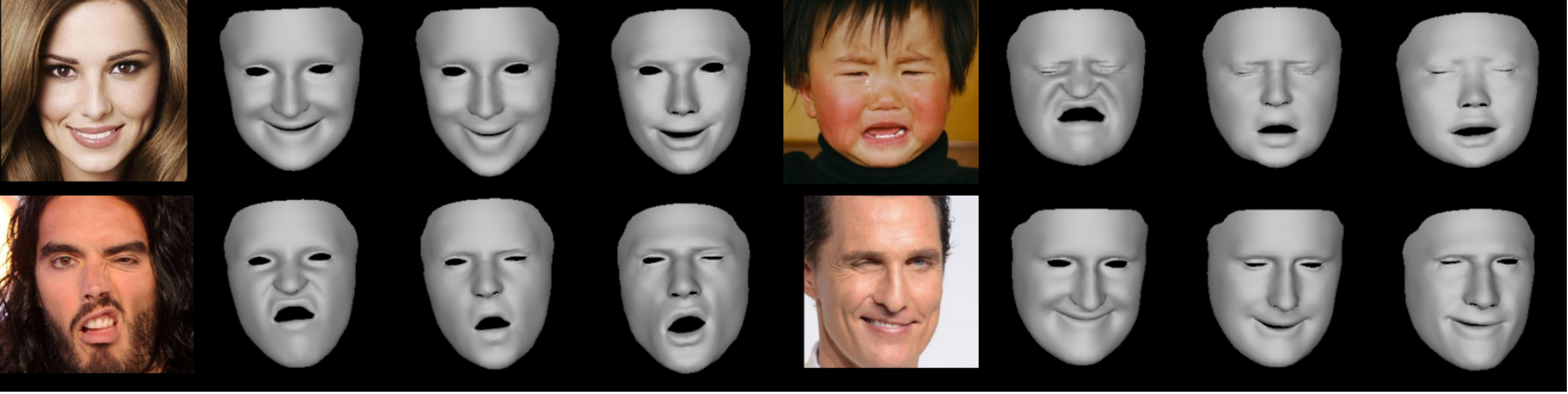}
    \caption{\textbf{3D reconstruction of emotions}. From left to right: input image, EMOCA v1, EMOCA v2, SMIRK. EMOCA v1 tends to exaggerate emotions, hence the highest score in emotion recognition. EMOCA v2, on the other hand, often lacks visual consistency with the original face, possibly due to a domain mismatch in the employed emotion recognition loss.}
    \label{fig:smirk_emoca_ex}
\end{figure}

In Figure \ref{fig:smirk_emoca_ex} we also show some qualitative examples comparing SMIRK, against EMOCAv1 and EMOCAv2. As it can be seen, EMOCAv1 which achieves the highest emotion recognition accuracy under this protocol tends to significantly exaggerate the observed emotion. On the other hand, EMOCAv2 often lacks the visual consistency with the original face. This could be attributed to the domain mismatch in the emotion recognition loss in EMOCA, since a textured rendered face with albedo is compared with the original image.


\begin{figure}[h]
    \centering
    \includegraphics[width=.6\linewidth]{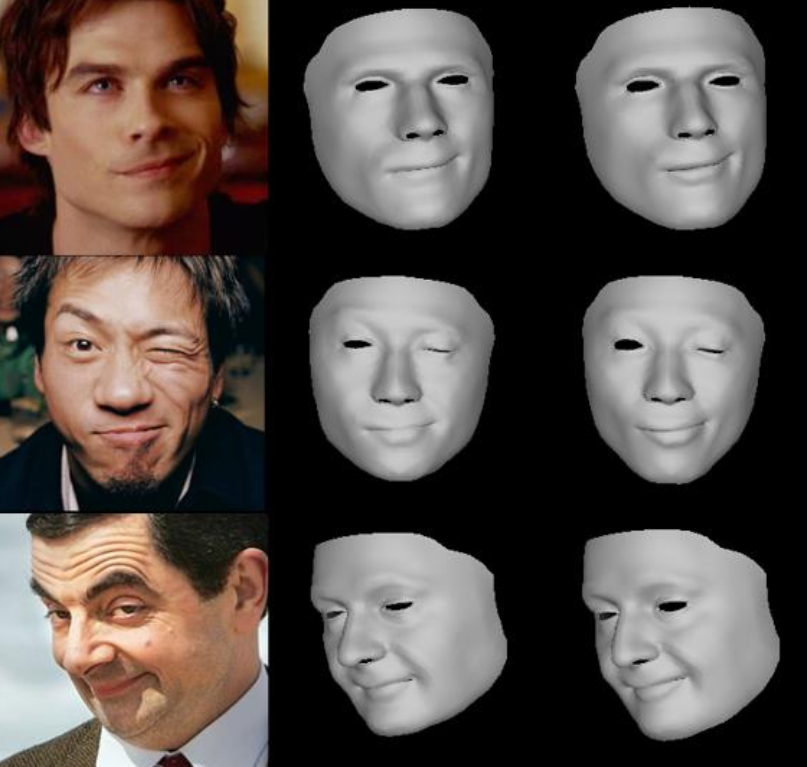}
    \caption{SMIRK with (middle column) and without (right column) pretraining the expression encoder achieves comparable results.}
    \label{fig:abl_pre}
\end{figure}

\subsection{Expression Pretraining Ablation}
We also evaluate the proposed pipeline, when the expression encoder is not initialized by the pre-trained network and show results in terms of 3D reconstruction in Table~\ref{tab:abl_pre} and qualitative in Figure~\ref{fig:abl_pre}. As we can see, training the expression encoder from scratch achieves similar and comparable results compared to using a pretrained expression encoder on landmarks only.


\begin{table}[]
\small
\centering
    \begin{tabular}{l|c|c|l}
        & mean $\downarrow$ & median $\downarrow$ & max $\downarrow$  \\ \toprule
 SMIRK w exp. pretrain  & 1.28 & 1.05 & 5.98 \\
SMIRK w/o exp. pretrain & 1.31 & 1.08 & 6.07 \\
\bottomrule
\end{tabular}
\caption{Ablation study on the effect of pretraining the expression encoder on the MultiFace dataset (per-vertex 3D reconstruction error in mm).}
\label{tab:abl_pre}
\end{table}

\section{Limitations \& Future Directions}
\label{sec:limitations}

\begin{figure}[h]
    \centering
    \includegraphics[width=1.0\linewidth]{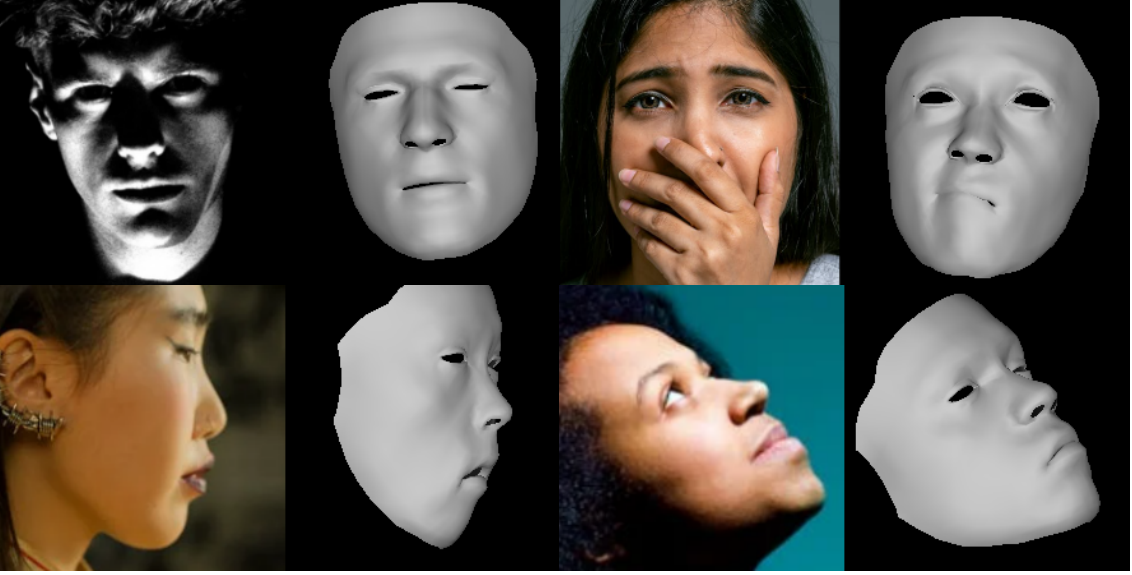}
    \caption{Examples where the SMIRK produces degraded results due to occlusions, extreme poses, and challenging lighting conditions.}
    \label{fig:limitations}
\end{figure}

Despite the effectiveness of the proposed method, there are specific limitations to be addressed, each one of them constituting a potential future direction:
\begin{itemize}
    \item \emph{Occlusions, Extreme Poses, and Challenging Lighting Conditions:} The majority of the datasets used in the proposed method have limited occluded cases and extreme poses. This makes the method sensitive to occlusions, as it tends to assume more intense expressions where a part is missing, rather than extrapolating from the existing information and retaining a more "average" expression for the missing parts. Additionally, the method can produce degraded results under cases with very limited lighting, as demonstrated in Figure~\ref{fig:limitations}. Nonetheless, addressing such cases was not within the scope of this work.

    \item \emph{Temporal Consistency:}
    the proposed framework has been trained on single images and the temporal aspect is not explored. 
    Smooth temporal transition and consistency can be imposed through external losses for video input.
    Towards this concept, one could extend the set of perceptual losses by adding a lip reading term, as in \cite{filntisis2023spectre}.
    \item \emph{Extension to Shape/Identity Parameters:}
    The present work focuses on estimating expression parameters, but the overall concept of learning through Analysis-by-Neural-Synthesis can be straightforwardly extended to estimate pose or identity parameters. Nonetheless, preliminary experiments showed that we cannot successfully optimize these parameters all-together, without sacrificing performance. Changing pose and shape each iteration affects the expression performance, not letting the expression parameters to capture finer subtle expressions due to ``jittering" effects of continuously changing pose/identity.  
    Nonetheless, given a good pose and expression estimation, one could fine-tune the shape parameters etc.
    Of course, optimizing shape also requires an extra set of regularization losses (e.g., shape consistency between different pictures of the same person).
\end{itemize}

\section{Additional Qualitative Results}
\label{sec:add_qual}

To further understand the effectiveness of SMIRK, we present a large set of visual examples in Figure~\ref{fig:visual_examples}, where our method is compared against other state-of-the-art approaches.
\begin{figure*}[t]
    \centering
    \includegraphics[width=.9\linewidth]{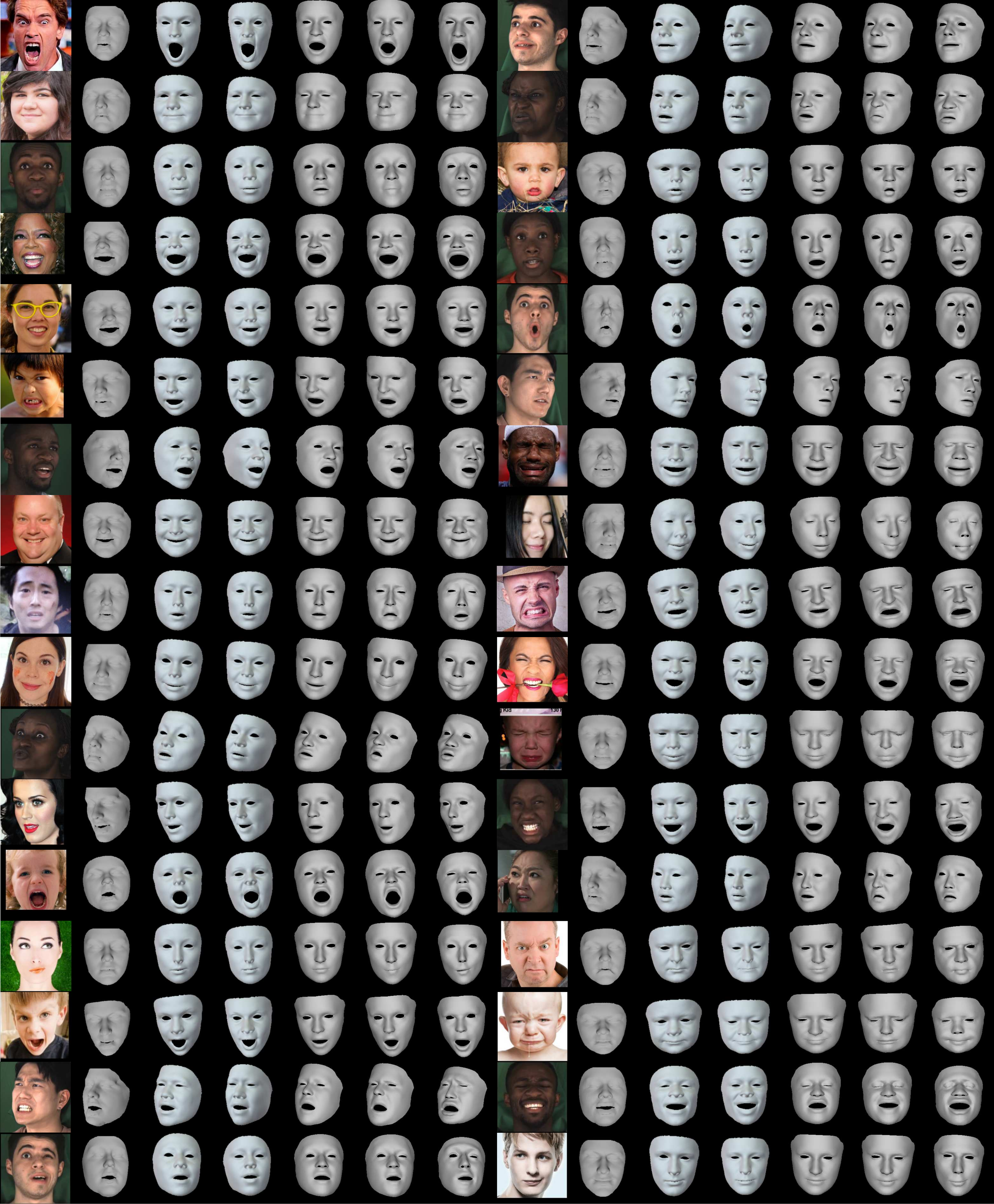}
    \caption{More qualitative results and visual comparisons of 3D face reconstruction from our method and four others. From left to right: Input, LeMoMo\cite{tewari2021learning} (method results provided by the authors), Deep3DFaceRecon(\cite{deng2019accurate}, FOCUS\cite{li2021fit}, DECA\cite{feng2021learning}, EMOCAv2\cite{danecek2022emoca}, and SMIRK. Please zoom in for details. Video results can also be found in the supplementary video.}
    \label{fig:visual_examples}
\end{figure*}
\clearpage
\clearpage


\end{document}